\documentclass{article}

\usepackage{booktabs}
\usepackage{bbm}
\usepackage{svg}
\usepackage[inline]{enumitem}
\usepackage{amsmath}
\usepackage{cancel}
\usepackage{changepage}

\usepackage{centernot}
\usepackage{diagbox}
\usepackage{subcaption}
\usepackage{wrapfig}
\usepackage{makecell}

\newcommand{\loss}{\mathcal{L}}
\newcommand{\lossfunc}{\mathcal{L}^{\text{point}}}
\newcommand{\losssample}{\mathcal{L}^{\text{dist}}}
\newcommand{\losslit}{\loss^\text{literature}}
\newcommand{\V}{\mathbb{V}}

\usepackage{tikz}

\def\imagewidth{0.7}
\def\numdata{n}

\newcommand{\psa}{P^*_A}
\newcommand{\pta}{P^\theta_A}
\newcommand{\psy}{P^{*}_{y|x}}
\newcommand{\ptpy}{P^{\theta,\psi}_{y|x}}

\newcommand{\E}{\mathbb{E}}
\newcommand{\R}{\mathbb{R}}

\definecolor{citeblue}{RGB}{0,0,180} 
\definecolor{citered}{RGB}{180,0,0} 

\definecolor{orange}{RGB}{240,140,0}
\definecolor{blue}{RGB}{40, 40, 200}
\definecolor{ggreen}{RGB}{0,150,100}
\definecolor{rred}{RGB}{180,0,40}

\usepackage{microtype}
\usepackage{graphicx}
\usepackage{booktabs} 

\usepackage{hyperref}

\usepackage[accepted]{icml2025}

\usepackage{amsmath}
\usepackage{amssymb}
\usepackage{mathtools}
\usepackage{amsthm}

\usepackage[capitalize,noabbrev]{cleveref}

\theoremstyle{plain}
\newtheorem{theorem}{Theorem}[section]
\newtheorem{proposition}[theorem]{Proposition}
\newtheorem{lemma}[theorem]{Lemma}
\newtheorem{corollary}[theorem]{Corollary}
\theoremstyle{definition}

\newtheorem{assumption}[theorem]{Assumption}
\theoremstyle{remark}

\usepackage[textsize=tiny]{todonotes}

\begin{document}

\twocolumn[
\icmltitle{Learning Latent Graph Structures and their Uncertainty}

\icmlsetsymbol{equal}{*}

\begin{icmlauthorlist}
\icmlauthor{Alessandro Manenti}{yyy}
\icmlauthor{Daniele Zambon}{yyy}
\icmlauthor{Cesare Alippi}{yyy,xxx}

\end{icmlauthorlist}

\icmlaffiliation{yyy}{Universit\`a della Svizzera italiana, IDSIA, Lugano, Switzerland}
\icmlaffiliation{xxx}{Politecnico di Milano, Milan, Italy}

\icmlcorrespondingauthor{Alessandro Manenti}{alessandro.manenti@usi.ch}

\icmlkeywords{Machine Learning, ICML}

\vskip 0.3in
]

\printAffiliationsAndNotice{}  

\begin{abstract}
Graph neural networks use relational information as an inductive bias to enhance prediction performance. Not rarely, task-relevant relations are unknown and graph structure learning approaches have been proposed to learn them from data. Given their latent nature, no graph observations are available to provide a direct training signal to the learnable relations. Therefore, graph topologies are typically learned on the prediction task alongside the other graph neural network parameters.
In this paper, we demonstrate that minimizing point-prediction losses does not guarantee proper learning of the latent relational information and its associated uncertainty. Conversely, we prove that suitable loss functions on the stochastic model outputs simultaneously grant solving two tasks: (i) learning the unknown distribution of the latent graph and (ii) achieving optimal predictions of the target variable. 
Finally, we propose a sampling-based method that solves this joint learning task \footnote{Code available at \hyperlink{https://github.com/allemanenti/Learning-Calibrated-Structures}{https://github.com/allemanenti/Learning-Calibrated-Structures}}. Empirical results validate our theoretical claims and demonstrate the effectiveness of the proposed approach.
\end{abstract}

\section{Introduction}\label{sec: Introduction}
Relational information processing has provided breakthroughs in the analysis of rich and complex data coming from, e.g., social networks, natural language, and biology. This side information takes various forms, from structuring the data into clusters to defining causal relations and hierarchies, and enables machine learning models to condition their predictions on dependency-related observations. 
In this context, predictive models take the form $y=f_\psi(x, A)$, where the input-output relation $x \mapsto y$ -- modeled by $f_\psi$ and its parameters in $\psi$ -- is conditioned on the relational information encoded in variable $A$. 
Graph Neural Networks (GNNs)~\cite{scarselli2008graph} are one example of models of this kind that rely on a graph structure represented as an adjacency matrix $A$ and have been demonstrated successful in a plethora of applications \citep{fout2017protein,shlomi2020graph}. 

Indeed, relational information is needed to implement such a relational inductive bias and, in some cases, it is provided at the application design phase. However, more frequently, such topological information is not rich enough to address the problem at hand, and -- not seldom -- it is completely unavailable.
Therefore, Graph Structure Learning (GSL) emerges as an approach to learn the graph topology \cite{kipf2018neural,franceschi2019learning, yu2021graph, fatemi2021slaps, zhu2021deep, cini2023sparse} alongside the predictive model $f_\psi$. This entails formulating a joint learning process that learns the adjacency matrix $A$ -- or a parameterization of it -- along with the predictor's parameters $\psi$. This is usually achieved by optimizing a loss function on the model output $y$, e.g., a point prediction measure based on the square or the absolute prediction error.

Different sources of uncertainty affect the graph structure learning process, including epistemic uncertainty in the data and variability inherent in the data-generating process. 
Learning appropriate models of the data-generating process can provide valuable insights into the modeled environment with uncertainty quantification enhancing explainability and interpretability, ultimately enabling more informed decision-making. Examples of applications are found in the study of infection and information spreading, as well as biological systems \citep{gomez2013structure, lokhov2016reconstructing, deleu2022bayesian}.
It follows that a probabilistic framework is appropriate to accurately capture the uncertainty in the learned relations whenever randomness affects the graph topology.
Probabilistic approaches have been devised in recent years. For instance, research carried out by \citet{franceschi2019learning,zhang2019bayesian,elinas2020variational,cini2023sparse} propose methods that learn a parametric distribution $P^\theta_A$ over the latent graph structure $A$. However, none of them have studied whether these approaches were able to learn a \emph{calibrated} latent distribution $\pta$ for $A$, properly reflecting the uncertainty associated with the learned topology.

In this paper, we fill this gap by addressing the joint problem of learning a predictive model yielding optimal point-prediction performance of the output $y$ and, contextually, a calibrated distribution for the latent adjacency matrix $A$. In particular, the novel contributions can be summarized as:
    \begin{enumerate}
        \item We demonstrate that models trained to achieve optimal point predictions do \emph{not} guarantee calibration of the adjacency matrix distribution [Section~\ref{sec: Optimizing Point Predictions}].
        \item We provide theoretical conditions on the predictive model and loss function that guarantee simultaneous calibration of the latent variable and optimal point predictions [Section~\ref{sec: Optimizing Distribution Discrepancy losses}].
        \item We propose a theoretically grounded sampling-based learning method to address the joint learning problem 
         [Section~\ref{sec: Optimizing Distribution Discrepancy losses}].
        \item We empirically validate the theoretical developments and claims presented in this paper and show that the proposed approach outperforms existing methods in solving the joint learning task [Section~\ref{sec: Experiments}].
    \end{enumerate}

Finally, we emphasize the significance of our contribution. 
The inherent latent nature of $A$ presents substantial learning challenges. Real-world applications rarely provide direct observations of the (latent) graph structure, making it impossible to use such data as learning signals for training the graph distribution $P_A^\theta$. This lack of real-world observations not only hampers model training but also complicates empirical evaluation of the learned latent distribution. Consequently, flawed decisions may be derived from learned models. 
This work addresses these limitations by (i) establishing theoretical guarantees for more robust learning of the latent variable to mitigate the need for evaluation on real data, and (ii) conducting a rigorous empirical analysis on synthetic datasets that provide the -- otherwise missing -- ground-truth knowledge required for an accurate validation of our claims.

\section{Related Work} 
\label{sec: Related Works}

\paragraph{Graph Structure Learning}
GSL is often employed end-to-end with a predictive model to better solve a downstream task.
Examples include applications within graph deep learning methods for static \cite{jiang2019semi, yu2021graph, kazi2022differentiable} and temporal data \cite{wu2019graph, wu2020connecting, cini2023sparse, defelice2024graphbased}; a recent review is provided by \citet{zhu2021deep}. 
Some approaches from the literature model the latent graph structure as stochastic \cite{kipf2018neural, franceschi2019learning, elinas2020variational, shang2021discrete, cini2023sparse}, mainly as a way to enforce sparsity of the adjacency matrix. 
To operate on discrete latent random variables, \citet{franceschi2019learning} utilize straight-through gradient estimations, \citet{cini2023sparse} rely on score-based gradient estimators, while \citet{niepert2021implicit} design an implicit maximum likelihood estimation strategy. 
A different line of research is rooted in graph signal processing, where the graph is estimated from a constrained optimization problem and the smoothness assumption of the signals \citep{kalofolias2016learn, dong2016learning, mateos2019connecting, coutino2020state, pu2021learning}.
A few works from the Bayesian literature have tackled the task of estimating uncertainties associated with graph edges. The model-based approaches by \citet{lokhov2016reconstructing} and \citet{gray2020bayesian} are two examples tackling relevant applications benefiting from uncertainty quantification.
Within the deep learning literature, \citet{zhang2019bayesian} propose a Bayesian Neural Network (BNN) modeling the random graph realizations. 
Differently, \citet{wasserman2024graph} develop an interpretable BNN designed over graph signal processing principles using unrolled dual proximal gradient iterations.
While some results on the output calibration are exhibited, to the best of our knowledge, no guarantee or evidence of calibration of the latent variable is provided, which we study in this paper instead.

\paragraph{Calibration of the model's output}
Research on model calibration has primarily focused on obtaining accurate and consistent predictions of the statistical properties of the target (random) variables $y$, from which uncertainty estimates on the model's predictions are derived.
For discrete outputs, such as in classification tasks, \citet{guo2017calibration} investigated the calibration of modern deep learning models and proposed temperature scaling as a solution. Other techniques in the same context include Histogram Binning \citep{zadrozny2001obtaining}, Cross Entropy loss with label smoothing \citep{muller2019does_lable_smoothing}, and Focal Loss \citep{mukhoti2020calibrating_focal_loss}. For continuous output distributions, \citet{laves2020well} proposed $\sigma$ scaling, while \citet{kuleshov2018accurate} developed a technique inspired by Platt scaling. More recently, conformal prediction techniques \citep{shafer2008tutorial} have gained popularity for providing confidence intervals in predictions.
We stress that within this paper, we are mainly concerned with latent variable calibration, rather than output calibration, although the two are related to each other.

\paragraph{Deep latent variable models}
Latent variables are extensively used in deep generative modeling \citep{kingma2013auto, rezende2014stochastic}, both with continuous and discrete latent variables \citep{van2017neural, bartler2019training}. In deep models, latent random variables often lack direct physical meaning, with only the outputs being collected for training. Therefore, studies mainly focused on maximizing the likelihood of the observed outputs in the training set, rather than calibrating the latent distribution. A few works proposed regularization of the latent space to improve stability and accuracy \citep{xu2018spherical, joo2020dirichlet}, facilitate smoother transitions in the output when the latent variable is slightly modified \citep{hadjeres2017glsr}, and apply other techniques aimed at enhancing data generation or improving model performance in general \citep{connor2021variational}.

To the best of our knowledge, no prior work has studied the joint learning problem of calibrating the latent graph distribution while achieving optimal point predictions.

\section{Problem Formulation}\label{sec: Problem Formulation}

Consider a set  of $N$ interacting entities and the data-generating process  
\begin{equation}
\label{eq:system-model} 
\begin{cases}
    A \sim P^*_A \\
    y = f^*(x, A)
\end{cases}
\end{equation}
where $y\in\mathcal Y$ is the system output obtained from input  $x\in\mathcal X$ through function $f^*$ and conditioned on a realization of the latent adjacency matrix $A\in \mathcal A\subseteq \{0,1\}^{N\times N}$ drawn from distribution $P^*_A$; input $x$ is assumed to be drawn from any distribution $P_x^*$ and superscript $*$ refers to unknown entities.
Each entry of the adjacency matrix $A$ is a binary value encoding the existence of a pairwise relation between two nodes. 
In the sequel, $x\in\mathcal X\subseteq  \R^{N\times d_{in}}$ and $y\in\mathcal Y\subseteq  \R^{N\times d_{out}}$ are stacks of $N$ node-level feature vectors of dimension $d_{in}$ and $d_{out}$, respectively, representing continuous inputs and outputs.

Given a training dataset $\mathcal D=\{(x_i,y_i)\}_{i=1}^\numdata$ of $\numdata$ input-output observations from \eqref{eq:system-model}, we aim at learning a probabilistic predictive model 
\begin{equation}
\label{eq:approx-model} 
\begin{cases}
    A \sim P_A^\theta \\
    \hat{y} = f_\psi(x, A)
\end{cases}
\end{equation}
from  $\mathcal D$, while learning at the same time distribution $P_A^\theta$ approximating  $P_A^*$.
The two parameter vectors $\theta$ and $\psi$ are trained to approximate distinct entities in \eqref{eq:system-model}, namely the distribution $P^*_A$ and function $f^*$, respectively. We assume 
\begin{assumption}\label{a:inclusion=in-family}
The family $\{P_A^\theta\}$ of probability distributions $P_A^\theta$ parametrized by $\theta$ and the family of predictive functions $\{f_\psi\}$ are expressive enough to contain the true latent distribution $P^*_A$ and function $f^*$, respectively.
\end{assumption}
Assumption \ref{a:inclusion=in-family} implies that $f^*\in\{f_\psi\}$ and $P_A^*\in\{P_A^\theta\}$ but does not request uniqueness of the parameters vectors $\psi^*$ and $\theta^*$ such that $f_{\psi^*}=f^*$ and $P_A^{\theta^*}=P_A^*$. Under such assumption the minimum function approximation error 
is null and we can focus on the theoretical conditions requested to guarantee successful learning, i.e., achieving both optimal point predictions and latent distribution calibration. In Section~\ref{sec: perturbed f}, we empirically show that the theoretical results can extend beyond this assumption in practice.

\paragraph{Optimal point predictions}
\label{sec:optimization objective - optimal point predictions}

Outputs $y$ and $\hat{y}$ of probabilistic model \eqref{eq:system-model} and \eqref{eq:approx-model} are random variables following push-forward distributions
    \footnote{The distribution of $y=f^*(x, A)$ originated from $P_A^*$ and of $\hat y=f_\psi(x,A)$ originated from $\pta$.}
$\psy$ and $P_{y|x}^{\theta,\psi}$, respectively.  A single point prediction $y_{PP}\in\mathcal Y$ can be obtained through an appropriate functional $T[\cdot]$ as
\begin{equation}\label{eq:single point prediction using T definition}
y_{PP}=y_{PP}(x,\theta,\psi)\equiv T\left[P_{y|x}^{\theta,\psi}\right].
\end{equation}
For example, $T$ can be the expected value or the value at a specific quantile. We then define an \textit{optimal predictor} as one whose parameters $\theta$ and $\psi$ minimize the expected \textit{point-prediction loss}
\begin{equation}\label{eq:T-bayes-rules}
\loss^{point}(\theta,\psi) = \mathbb E_{x\sim P_x^*}\left[\mathbb E_{y\sim P^*_{y|x}}\left[ \ell\big(y, y_{PP}(x,\theta,\psi)\big) \right]\right]
\end{equation}
between the system output $y$ and the point-prediction $y_{PP}$, as measured by of a loss function $\ell:\mathcal Y\times \mathcal Y\to \mathbb R_+$.

Statistical functional $T$ is coupled with the loss $\ell$ as the optimal functional $T$ to employ given a specific loss $\ell$ is often known  \cite{berger1990statistical, gneiting2011making}, when $P_{y|x}^{\theta,\psi}$ approximates well $P_{y|x}^*$.
For instance, if $\ell$ is the Mean Absolute Error (MAE) the associated functional $T$ is the median, if $\ell$ is the Mean Squared Error (MSE) the associated functional is the expected value.

\paragraph{Latent distribution calibration}
\label{sec:optimization objective - latent distribution calibration}
 Calibration of a parametrized distribution {$\pta$} requires learning parameters $\theta$, so that $P^\theta_A$ aligns with {true distribution} $P^*_A$. Quantitatively, a dissimilarity measure 
$
\Delta^{cal} : \mathcal P_A\times\mathcal P_A \to \mathbb R_+,
$
defined over a set $\mathcal P_A$ of distributions on $\mathcal A$, assesses how close two distributions are. The family of $f$-divergences \cite{renyi1961measures}, such as the Kullback-Leibler divergence, and the integral probability metrics \cite{muller1997integral}, such as the maximum mean discrepancy \cite{gretton2012kernel} are examples of such dissimilarity measures. In this paper, we are interested in those discrepancies for which 
$
\Delta^{cal}(P_1, P_2) = 0 \iff P_1 = P_2
$ holds. 
It follows that the latent distribution $\pta$ is \textit{calibrated} on $\psa$ if it minimizes the \textit{latent distribution loss}
\begin{equation}\label{eq:calibration-loss}
\mathcal L^{cal}= \mathbb E_{x\sim P_x^*} \left[\Delta^{cal}\left(P^*_{A}, P^\theta_{A}\right)\right],
\end{equation}
or simply $\mathcal L^{cal}=\Delta^{cal}\left(P^*_{A}, P^\theta_{A}\right)$, when $A$ and $x$ are independent.

The problem of designing a predictive model \eqref{eq:approx-model} that both yields {optimal point predictions} (i.e., minimizes $\mathcal L^{point}$ in \eqref{eq:T-bayes-rules}) and  {calibrates the latent distribution} (i.e., minimizes $\mathcal L^{cal}$ in \eqref{eq:calibration-loss}) is non-trivial for two main reasons. At first, as the latent distribution $P_A^*$ is unknown (and no samples from it are available), we cannot directly estimate $\mathcal L^{cal}$. 
Second, as shown in Section~\ref{sec: Optimizing Point Predictions}, multiple sets of $\theta$ parameters may minimize $\loss^{point}$ without minimizing $\loss^{cal}$.

\section{Limitations of Point-Prediction Optimization}\label{sec: Optimizing Point Predictions}
\label{sec:optimizing point predictions}

In this section, we demonstrate that the optimization of a point prediction loss  (Equation~\eqref{eq:T-bayes-rules}) does not generally grant calibration of the latent random variable $A$. 
\begin{proposition}
\label{th:optimal prediction don't guarantee calibration}
    Consider Assumption~\ref{a:inclusion=in-family}.
    Loss function $\loss^{point}(\theta,\psi)$ in \eqref{eq:T-bayes-rules} is minimized by all $\theta$ and $\psi$ s.t.\ $T\Big[P_{y|x}^{\theta,\psi}\Big] = T\Big[P^*_{y|x}\Big]$ almost surely on $x$ and, in particular,
    $$
    \loss^{point}(\theta,\psi) \text{ is minimal}
    \quad
    \begin{array}{cc}
    \Longleftarrow
    \\
    \;\;\not\!\!\Longrightarrow
    \end{array}
    \quad
    P_{y|x}^{\theta,\psi} = P^*_{y|x}.
    $$
\end{proposition}

The proof of the proposition is given in Appendix \ref{appendix: optimal prediction don't guarantee calibration}; we provide a counterexample for which calibration is not granted even when the processing function $f_\psi$ is equal to $f^*$ in Appendix \ref{appendix MAE and MSE no calibration}.

The limitation of point-prediction losses is also empirically demonstrated in Section~\ref{sec: comparison with literature}, Table~\ref{table: empirical calibration}, where it is shown that optimizing point-prediction losses does not grant calibration

Given the provided negative result and the impossibility of assessing loss $\mathcal L^{cal}$ in \eqref{eq:calibration-loss}, in the next section, we propose another optimization objective that, as we will prove, allows us to both calibrate the latent random variable and to have optimal point predictions.

\section{Predictive Distribution Optimization: Two Birds with One Stone}
\label{sec: Optimizing Distribution Discrepancy losses}
\label{sec:calibrating latent distribution}

In this section, we show that we can achieve an optimal point predictor \eqref{eq:approx-model} and a calibrated latent distribution $P_A^\theta$ by comparing
push-forward distributions $P_{y|x}^*$ and $P_{y|x}^{\theta,\psi}$ of the outputs $y$ conditioned on input $x$. In particular, Theorem~\ref{theo:Ldist-Lpoint-Lcal} below proves that, under appropriate conditions, minimization of the \emph{output distribution loss}
\begin{equation}\label{eq:distribution discrepancy loss}
    \mathcal L ^{dist}(\theta,\psi)=\mathbb E_{x\sim P_x^*}\Big[ \Delta(P_{y|x}^*,P_{y|x}^{\theta,\psi}) \Big]
\end{equation}
provides calibrated $P_A^\theta$, even when $P_A^*$ is not available; $\Delta:\mathcal P_y\times \mathcal P_y\to \mathbb R_+$ is a dissimilarity measure between distributions over space $\mathcal Y$. We assume the following on dissimilarity measure $\Delta$.
\begin{assumption}\label{a:Delta-premetric}
     $\Delta(P_1,P_2)$ $\geq 0$ for all distributions $P_1$ and $P_2$ in $\mathcal P_y$ and $\Delta(P_1,P_2) = 0$ if and only if $ P_1=P_2$. 
\end{assumption}
Several choices of $\Delta$ meet Assumption~\ref{a:Delta-premetric}, e.g., $f$-divergences and some integral probability metrics \cite{muller1997integral}; the dissimilarity measure $\Delta$ employed in this paper is discussed in Section~\ref{sec:mmd}.

\begin{theorem}\label{theo:Ldist-Lpoint-Lcal}
Let $I=\{x:A\mapsto f^*(x,A) \text{ \normalfont is injective}\}\subseteq \mathcal X$ be the set of points $x\in\mathcal X$ such that map $A\mapsto f^*(x,A)$ is injective. 
Under Assumptions~\ref{a:inclusion=in-family} and \ref{a:Delta-premetric}, 
if $\mathbb P_{x\sim P_x^*}(I)>0$ and $\psi^*$ is such that $f_{\psi^*}=f^*$, then 
\begin{align*}
\mathcal L^{dist}(\theta,\psi^*) = 0 &\implies
\begin{cases}\mathcal L^{point}(\theta,\psi^*) \text{ is minimal}
\\
\mathcal L^{cal}(\theta) =0.
\end{cases}
\end{align*}
\end{theorem}
Theorem~\ref{theo:Ldist-Lpoint-Lcal} is proven in Appendix \ref{appendix: l dist for calibration pt 1}.
Under the theorem's hypotheses, a predictor that minimizes $\loss^{dist}$ is both \textit{calibrated} on the latent random distribution and provides \textit{optimal point predictions}. This overcomes limits of Proposition \ref{th:optimal prediction don't guarantee calibration} where optimization of $\loss^{point}(\theta,\psi^*)$ does not grant $\loss^{cal}(\theta)=0$.

The hypotheses under which Theorem~\ref{theo:Ldist-Lpoint-Lcal} holds are rather mild. In fact, condition $\mathbb P_{x\sim P_x^*}(I)>0$ pertains to the data-generating process and intuitively ensures that, for some $x$, different latent random variables produce different outputs.
A sufficient condition for $\mathbb P_{x\sim P_x^*}(I)>0$ to hold is the existence of a point $\bar x$ in the support of $P_x^*$ such that $A\mapsto f^*(\bar x, A)$ is injective with $f^*$ continuous w.r.t.\ $\bar x$; see Corollary \ref{corollary: l dist for calibration pt 2} in Appendix~\ref{appendix: l dist for calibration pt 2}. Although only a single point $\bar{x}$ is required, having more points that satisfy the condition simplifies the training of the parameters. Corollary \ref{corollary: l dist for calibration pt 2} holds for arbitrarily complex processing functions $f^*$. 
More specifically, when considering simple GNN layers and discrete latent matrices $A$, we can prove that the condition $\mathbb P_{x\sim P_x^*}(I)>0$ is $-$ except from pathological cases $-$ always satisfied (see Proposition \ref{prop: l dist for calibration pt 2} in Appendix~\ref{appendix: l dist for calibration pt 2}). 
Instead, condition $f_\psi=f^*$ is set to avoid scenarios of different, yet equivalent,
    \footnote{E.g., $f_\psi(A,x) = f_*(\mathbf 1-A,x)$ and $\pta$ encoding the absence of edges instead of their presence as in $\psa$.}
representations of the latent distribution. An empirical analysis of the theorem's assumptions is provided in Section~\ref{sec: perturbed f}, demonstrating that the theoretical results hold in practice, even when those assumptions do not strictly apply.

Assumptions~\ref{a:inclusion=in-family} and \ref{a:Delta-premetric} can be met with an appropriate choice of model \eqref{eq:approx-model} and measure $\Delta$; as such they are controllable by the designer.
Assumption~\ref{a:Delta-premetric} prevents from obtaining mismatched output distributions when $\mathcal L^{dist}(\theta,\psi)=0$ and can be easily satisfied.
As mentioned above, popular measures, e.g., the Kullback-Leibler divergence, meet the theorem's assumptions and therefore can be adopted as $\Delta$. However, as $f$-divergences rely on the explicit evaluation of the likelihood of $y$, they are not always practical to compute \cite{mohamed2016learning}. For this reason, we propose considering the Maximum Mean Discrepancy (MMD) \cite{gretton2012kernel} as a versatile alternative that allows Monte Carlo computation without requiring evaluations of the likelihood w.r.t.\ the output distributions $\psy$ and $\ptpy$. Energy distances \cite{szekely2013energy} provide an alternative feasible choice.

\subsection{Maximum Mean Discrepancy}\label{sec:mmd}

Given two distributions $P_1,P_2\in\mathcal P_y$, the MMD can be defined as 
\begin{equation}\label{eq: MMD definition}
    \text{MMD}_\mathcal{G}[P_1, P_2] = \sup_{g\in\mathcal G}\left\{\mathbb E_{y\sim P_1}\big[g(y)\big]-\mathbb E_{y\sim P_2}\big[g(y)\big]\right\},
\end{equation}
i.e., the supremum, taken over a set $\mathcal G$ of functions $\mathcal Y\to \mathbb R$, of the difference
between expected values w.r.t.\ $P_1$ and $P_2$.
An equivalent form is derived for a generic kernel function $\kappa(\cdot, \cdot):\mathcal Y \times \mathcal Y\to \mathbb R$:
\begin{multline}\label{eq: MMD^2 monte carlo friendly definition}
    \text{MMD}^2_{\mathcal{G}_\kappa}[P_1, P_2] = 
    \underset{y_1, y'_1 \sim P_1}{\E} \Big[\kappa(y_1 , y'_1) \Big]  \\
     -2 \underset{\substack{y_1 \sim P_1 \\ y_2 \sim P_2}}{\E} \Big[\kappa(y_1 , y_2) \Big] 
    + \underset{y_2, y'_2 \sim P_2}{\E} \Big[\kappa(y_2 , y'_2) \Big],
\end{multline}
and it is associated with the unit-ball $\mathcal G_k$ of functions in the reproducing kernel Hilbert space of $\kappa$; note that \eqref{eq: MMD^2 monte carlo friendly definition} is the square of \eqref{eq: MMD definition}. Moreover, when universal kernels are considered (e.g., the Gaussian one), then \eqref{eq: MMD^2 monte carlo friendly definition} fulfills Assumption~\ref{a:Delta-premetric} (see Theorem 5 in \citep{gretton2012kernel}).
Dissimilarity in \eqref{eq: MMD^2 monte carlo friendly definition} can be conveniently estimated via Monte Carlo (MC) and employed within a deep learning framework. Accordingly, we set $\Delta=\text{MMD}^2_{\mathcal{G}_\kappa}$ and learn parameter vectors $\psi$ and $\theta$ by minimizing $\mathcal L^{dist}(\theta,\psi)$ via gradient-descent methods.

\subsection{Finite-Sample Computation of the Loss}\label{sec: Practical Implementation}

To compute the gradient of $ \loss^{dist}(\theta,\psi)=\E_{x\sim P_x^*} \left[\text{MMD}^2_\mathcal{G_\kappa}\left[ P_{y|x}^{\theta,\psi},P_{y|x}^*\right]\right]$ w.r.t. parameter vectors $\psi$ and $\theta$, 
we rely on MC sampling to estimate in \eqref{eq:distribution discrepancy loss} 
expectations over input $x\sim P_x^*$, target output $y\sim P_{y|x}^*$ and model output $\hat y\sim P_{y|x}^{\theta,\psi}$ . This amounts to substitute  $\text{MMD}^2_\mathcal{G_\kappa}$ with
\begin{equation}\label{eq: MMD^2 monte carlo}
    \widehat{\text{MMD}}^2_{\theta,\psi}(x,y) =
    2\frac{\sum_{j<i=1}^{N_{adj}} \kappa(\hat y_i, \hat y_j)}{N_{adj}(N_{adj}-1)}
    -2 \frac{\sum_{i=1}^{N_{adj}} \kappa(y,\hat y_i)}{N_{adj}}.
\end{equation}
In \eqref{eq: MMD^2 monte carlo}, $N_{adj}>1$ is the number of adjacency matrices sampled from $P_A^\theta$ to obtain output samples $\hat y_i=f_\psi(x,A_i)\sim P_{y|x}^{\theta,\psi}$, whereas the pair $(x,y)$ is a pair from the training set $\mathcal D$. We remark that in \eqref{eq: MMD^2 monte carlo} the third term of \eqref{eq: MMD^2 monte carlo friendly definition} -- i.e., the one associated with the double expectation with respect to $P_{y|x}^*$ -- is neglected as it does not depend on $\psi$ and $\theta$. 

Gradient $\nabla_{\psi} \loss^{dist}(\theta,\psi)$ is computed via automatic differentiation  by averaging $\nabla_\psi \widehat{\text{MMD}}^2(\theta,\psi)$ within a mini-batch of observed data pairs $(x_i,y_i)\in\mathcal D$.
For $\nabla_\theta \loss^{dist}(\theta,\psi)$, the same approach is not feasible. This limitation arises because the gradient is computed with respect to the same parameter vector $\theta$ that defines the integrated distribution. 
Here, we rely on a score-function gradient estimator (SFE) \cite{williams1992simple, mohamed2020monte} which uses the log derivative trick to rewrite the gradient of an expected loss $L$ as
$\nabla_\theta\mathbb{E}_{A \sim P^\theta}[L(A)] = \mathbb{E}_{A \sim P^\theta}[L(A)\nabla_\theta\log P^\theta(A)]$, with $P^\theta(A)$ denoting the likelihood of $A\sim P^\theta$. 
Applying the SFE to our problem the gradient w.r.t.\ $\theta$ reads:
\begin{multline}\label{eq: d loss / d theta}
\nabla_\theta \loss^{dist} = 
\underset{x,y^*}{\E} \bigg[ \\
    \underset{\hat{y}_1, \hat{y}_2}{\E} \left[
        \kappa(\hat{y}_1, \hat{y}_2) 
        \nabla_\theta \log\left( P_{y|x}^{\theta,\psi}(\hat{y}_1)\ptpy(\hat{y}_2) \right) 
    \right] 
    \\
     -2 \underset{\hat{y}}{\E} \left[  \kappa(y^* , \hat{y})\nabla_\theta \log \ptpy(\hat{y}) \right]
    \bigg]
\end{multline}
where $\hat{y}_1,\hat{y}_2, \hat{y} \sim \ptpy$.
An apparent setback of SFEs is their high variance \cite{mohamed2020monte}, which we address in Section \ref{sec:variance-reduction} by deriving a variance-reduction technique based on control variates that requires negligible computational overhead.

\subsection{Variance-Reduced Loss for SFE}\label{sec:variance-reduction}
Two natural approaches to reduce the variance of MC estimates of \eqref{eq: d loss / d theta} involve (i) increasing the number $B$ of training data points in the mini-batch used for each gradient estimate and (ii) increasing the number $N_{adj}$ of adjacency matrices sampled for each data point in \eqref{eq: MMD^2 monte carlo}. 
These techniques act on two different sources of noise. Increasing $B$ decreases the variance coming from the data-generating process, whereas increasing $N_{adj}$ improves the approximation of the predictive distribution $P_{y|x}^{\theta,\psi}$. 
Nonetheless, by fixing $B$ and $N_{adj}$, it is possible to further reduce the latter source of variance by employing the \emph{control variates} method \cite{mohamed2020monte} that, in our case, requires only a negligible computational overhead but sensibly improves the training speed (see Section \ref{sec: Experiments}).

Consider the expectation $\E_{A \sim P^\theta} [ L(A) \nabla_\theta \log P^\theta(A)]$ of the SFE -- both terms in \eqref{eq: d loss / d theta} can be cast into that form. 
With the control variates method, a function with null expectation is subtracted from $L(A) \nabla_\theta \log P^\theta(A)$.
\begin{equation}\label{eq: control variate equation}
G(A) = L(A)\nabla_\theta \log P^\theta(A) - \beta \Big( h(A) - \E_{A \sim P^\theta} [ h(A)] \Big)
\end{equation}
that leads to a reduced variance in the MC estimator of the gradient while maintaining it unbiased. In this paper, we set function $h(A) $ to $\nabla_\theta \log P^\theta(A)$ and show how to compute a near-optimal choice for scalar value $\beta$, often called \emph{baseline} in the literature. As the expected value of $\nabla_\theta \log P^\theta(A)$ is zero, gradient \eqref{eq: d loss / d theta}  rewrites as
\begin{multline}\label{eq: d loss / d theta REDUCED VARIANCE}
\nabla_\theta \loss^{dist} \! = \hspace{-1mm} \underset{x,y^*}{\E} \bigg[ 
   -2 \underset{A}{\E} \! \left[  \left(\kappa(y^* , \hat{y})-\beta_2\right) \! \nabla_\theta \log \pta(A) \right] \\ 
    + \underset{A_{1} A_2}{\E} \Big[
        \left(\kappa(\hat{y}_1, \hat{y}_2)-\beta_1\right) 
        \;\nabla_\theta \log\left( \pta(A_1)\pta(A_2) \right) 
    \Big] \bigg]. 
\end{multline}
In Appendix \ref{appendix: optimal beta control variates}, we show that in our setup the best values of $\beta_1$ and $\beta_2$ are approximated by
\begin{align}\label{eq:optimal betas} \nonumber
    \tilde \beta_1 &= \E_{x}\Big[\E_{A_1, A_2\sim\pta} \Big[ \kappa\big(f_\psi(x ,A_1), f_\psi(x,A_2)  \big) \Big]\Big], \\
    \tilde \beta_2 &= \E_{x,y^*}\Big[\E_{A\sim\pta}\Big[ \kappa\big(y^*, f_\psi(x ,A) \big) \Big]\Big],
\end{align}
which can be efficiently computed via MC reusing the kernel values already computed for \eqref{eq: d loss / d theta REDUCED VARIANCE}.

\subsection{Computational Complexity}\label{sec: computational complexity}
Focusing on the most significant terms, for every data pair $(x,y)$ in the training set, computing the loss $\loss^{dist}$ requires $\mathcal{O}(N_{adj}^2)$ kernel evaluations $\kappa(\hat y_i,\hat y_j)$ in \eqref{eq: MMD^2 monte carlo}, $\mathcal{O}(N_{adj})$ forward passes through the GNN $\hat y_i = f_\psi(x,A_i)$ in \eqref{eq: MMD^2 monte carlo} and $\mathcal{O}(N_{adj})$ likelihood computations $\pta(A_i)$ in \eqref{eq: d loss / d theta REDUCED VARIANCE}. The computation of baselines $\beta_1$ and $\beta_2$ in \eqref{eq:optimal betas} requires virtually no overhead, as commented in previous Section~\ref{sec:variance-reduction}. 
Similarly, computing the loss gradients requires $\mathcal{O}(N_{adj}^2)$ derivatives for what concerns the kernels, $\mathcal{O}(N_{adj})$ gradients $\nabla_\psi \hat y_i$ and $\nabla_\theta \log \pta(A_i)$. We empirically observed that for $N_{adj} \ge 16$, both the latent distribution loss $\loss^{cal}$ and the point prediction loss $\loss^{point}$ of final models are equivalent for the considered problem. This suggests that $N_{adj}$ is not a critical hyperparameter.

Since we can employ sparse representations of adjacency matrices, the GNN processing costs scale linearly in the number of nodes $N$ for bounded-degree graphs. 
From our experience, the GNN processing is the most demanding operation and the cost of quadratic components, such as the parameterization of $\theta_{ij}$, do not pose significant overhead.

\section{Experiments}\label{sec: Experiments}

This section empirically validates the proposed technique and the main claims of the paper. 
While point predictions can be evaluated on observed input-output pairs $(x, y)$ provided as a test set, assessing latent-variable calibration performance -- the discrepancy between $\psa$ and the learned $\pta$ -- requires knowledge of the ground-truth latent distribution itself or of observation thereof.
Such ground-truth knowledge, however, is not available in real-world datasets, as the latent distribution is indeed unknown. Therefore, to validate the theoretical results, we designed a synthetic dataset that allows us to evaluate different performance metrics on both $y$ and $A$. We remark that the latent distribution is used \emph{only} to assess performance and does not drive the model training in any way.

Section~\ref{sec:exp:validation} demonstrates that the proposed approach can successfully solve the joint learning problem across different graph sizes and highlights the effectiveness of the proposed variance reduction technique. Section \ref{sec: perturbed f} empirically investigates the generality of the theoretical results we develop, demonstrating appropriate calibration of the latent distribution even in scenarios where the assumptions of Theorem \ref{theo:Ldist-Lpoint-Lcal} do not hold. Section \ref{sec: comparison with literature} demonstrates that the proposed approach is more effective than existing methods in solving the joint learning task. As a last experiment, in Appendix \ref{appendix: AQI experiment} we test our approach and show that sensible graph distributions can be learned in real-world settings.

\paragraph{Dataset and models} 
Consider data-generating process \eqref{eq:system-model} with latent distribution $P_A^*=P_A^{\theta^*}$ producing $N$-node adjacency matrices. 
Random graph $A\sim P_A^*$ is given as a set of independent edges $(i,j)$, for $i,j=1,\dots,N$, each of which is sampled with probability $\theta^*_{i,j}$. 
Function $f_{*}=f_{\psi^*}$ is a generic GNN with node-level readout, i.e., $f_{\psi^*}(\cdot, A):\R^{N\times d_{in}}\to \R^{N\times d_{out}}$. 
The components $\theta^*$ are set to either $0$ or $3/4$ according to the pattern depicted in Figure \ref{fig:stoc-gpvar-graph}; additional specifics are detailed in Appendix~\ref{appendix:experiments}. We result in a dataset of $35k$ input-output pairs $(x,y)$, 80\% of which are used as training set, 10\% as validation set, and the remaining 10\% as test set. As predictive model family \eqref{eq:approx-model}, we follow the same architecture of $f_{\psi^*}$ and $P_A^{\theta^*}$ ensuring that during all the experiments Assumption~\ref{a:inclusion=in-family} is fulfilled;
similar models have been used in the literature \citep{franceschi2019learning, elinas2020variational, kazi2022differentiable, cini2023sparse}. 
In Section \ref{sec: perturbed f} we test the method's validity beyond this assumption. The model parameters are trained by optimizing the expected squared MMD in \eqref{eq: MMD^2 monte carlo} with the rational quadratic kernel \cite{binkowski2018demystifying}.

\begin{figure}
  \centering
  \includegraphics[scale=0.75]{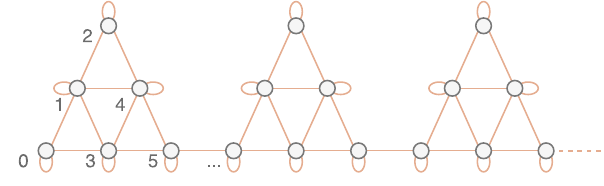}\\
  \includegraphics[scale=0.35]{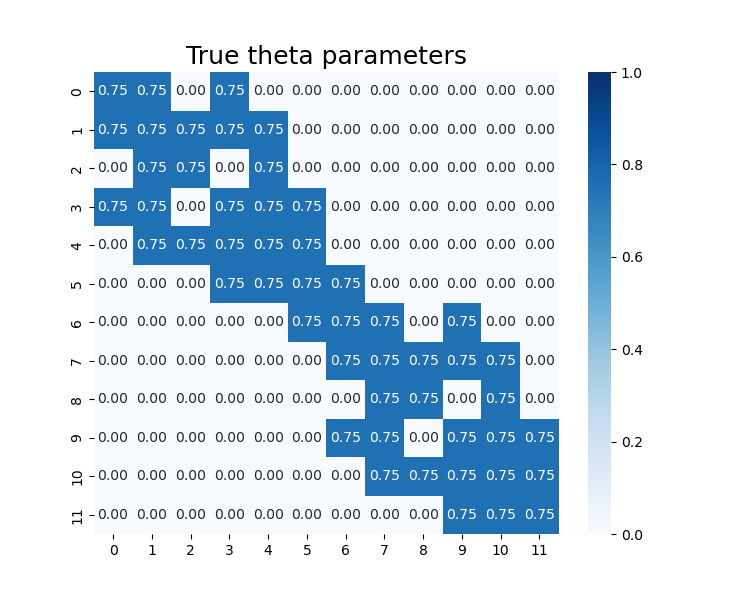}
  \caption{Adjacency matrices sampled from $P_A^*=P_A^{\theta^*}$ for the experiment of Section~\ref{sec: Experiments} are subgraphs of the top graph; in this picture, 3 communities of an arbitrarily large graph are shown.  Each edge in orange is independently sampled with probability $\theta^*_{ij}$; parameters $\theta^*_{ij}$ defining the edge probabilities are represented at the bottom for a two communities graph.}
  \label{fig:stoc-gpvar-graph}
\end{figure}

\subsection{Graph Structure Learning \& Optimal Point Predictions}\label{sec:exp:validation}
To test our method's ability to both calibrate the latent distribution and make optimal predictions, we train the model minimizing $\loss^{dist}$ as described in Section~\ref{sec: Practical Implementation}.

\begin{figure*}
\centering
    \begin{subfigure}[t]{0.3\linewidth}
    \centering
     \includegraphics[scale=0.285]{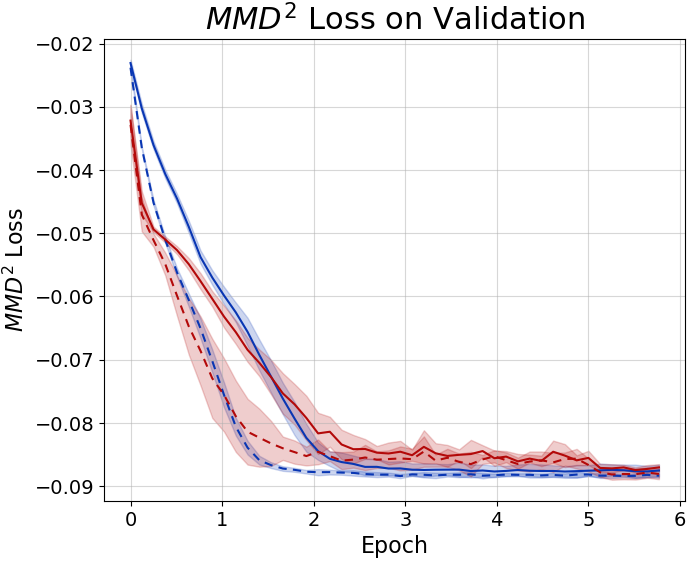}
     \caption{}
     \label{fig: MMD prediction loss}
\end{subfigure} \hfill
\begin{subfigure}[t]{0.3\linewidth}
     \centering
     \includegraphics[scale=0.285]{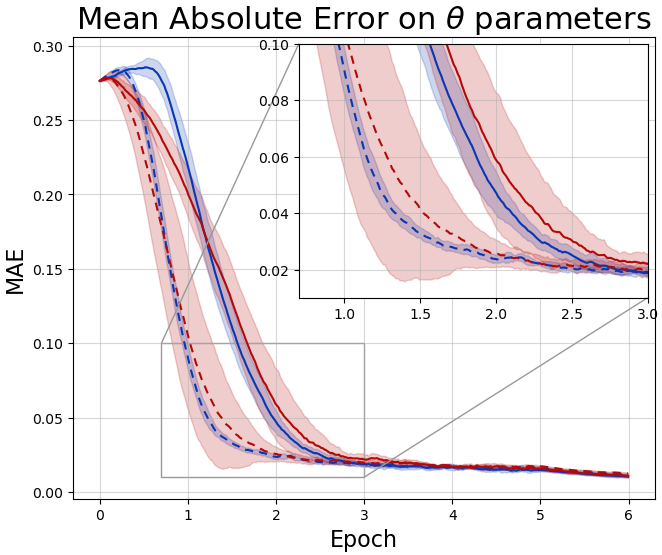}
     \caption{}
     \label{fig: MAE on theta}
\end{subfigure} \hfill
\begin{subfigure}[t]{0.3\linewidth}
     \centering
     \includegraphics[scale=.285]{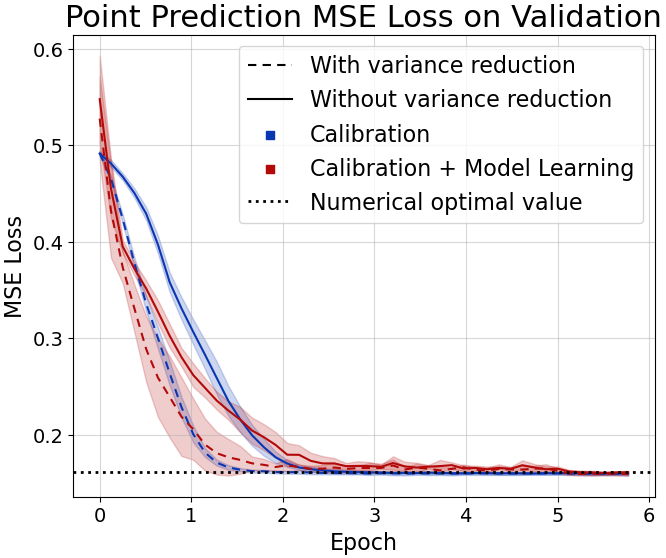}
     \caption{}
     \label{fig: MAE point predictions}
\end{subfigure}
\vspace{-.2cm}
\caption{Validation losses $\loss^{dist}$, $\loss^{cal}$ and $\loss^{point}$ during training. At epoch 5, the learning rate is decreased to ensure convergence. $\loss^{dist}$ in Subfigure \ref{fig: MMD prediction loss} is negative as the third term in \eqref{eq: MMD^2 monte carlo friendly definition} is constant and not considered.}
\label{fig:validation-losses}
\end{figure*}

Figure~\ref{fig:validation-losses} reports the validation losses during training: MMD loss $\loss^{dist}$ {as in \eqref{eq:distribution discrepancy loss}}, MAE between the learned parameters $\theta$ and the ground truth $\theta^*$ as $\loss^{cal}$ {\eqref{eq:calibration-loss}}, and point-prediction loss $\loss^{point}$ {as in \eqref{eq:T-bayes-rules}} with $\ell$ being the MSE. The results are averaged over 8 different model initializations and error bars report $\pm 1$ standard deviations from the mean. Results are reported with and without applying the variance reduction (Section~\ref{sec:variance-reduction}), by training only parameters $\theta$ while freezing $\psi$ to $\psi^*$ (same setting of Theorem~\ref{theo:Ldist-Lpoint-Lcal}), and by jointly training both $\psi$ and $\theta$.

\paragraph{Solving the joint learning problem} 
Figure~\ref{fig: MMD prediction loss} shows that the training succeeded and the MMD loss $\loss^{dist}$ converged to its minimum value.
\footnote{Numerical estimation shows that the minimum value of $\losssample$ for the given kernel is approximately $-0.088$; note that although the $\text{MMD}^2\ge 0$, the third term in \eqref{eq: MMD^2 monte carlo friendly definition} is dropped from \eqref{eq: MMD^2 monte carlo}.}
Having minimized $\loss^{dist}$, from Figure~\ref{fig: MAE on theta} we see that also the calibration of latent distribution $\pta$ was successful; in particular, the figure shows that the MAE on $\theta$ parameters ($N^{-2} \lVert \theta^* - \theta\rVert_1$) approaches zero as training proceeds (MAE $<0.01$). Regarding the point predictions, Figure~\ref{fig: MAE point predictions} confirms that $\loss^{point}$ reached its minimum value; recall that optimal prediction MSE is not $0$, as the target variable $y$ is random, and note that a learning rate reduction is applied at epoch number $5$.
The optimality of the point-prediction is supported also by the performance on separate test data and with respect to the MSE as point-prediction loss $\ell$. 
Moreover, we observe that calibration is achieved regardless of the variance reduction and whether or not parameters $\psi$ are trained.
Lastly, Figure \ref{fig: theta adj} in Appendix~\ref{appendix:additional-figs-validation} shows the learned parameters $\theta$ of the latent distribution and the corresponding absolute discrepancy resulted from a (randomly chosen) training run.

\paragraph{Variance reduction effectiveness}
Figures \ref{fig: MMD prediction loss}, \ref{fig: MAE on theta} and \ref{fig: MAE point predictions} demonstrate that the proposed variance reduction method (Section \ref{sec: Practical Implementation}) yields notable advantages training speed up (roughly $50\%$ faster). For this reason, the next experiments rely on variance reduction.

\paragraph{Larger graphs}\label{sec: larger graphs}
The theoretical results developed hold for any number of nodes $N$. However, the number of possible edges scales quadratically in the number of nodes -- a potential issue inherent to the GSL problem, not our approach. Therefore, for extremely large graphs the ratio between the number of free parameters in $\theta$ and the size of the training set can become prohibitive. 
Nonetheless, in Figure \ref{fig: larger graph N120}, we show all $\sim 15K$ parameters of the considered $\pta$ can be effectively learned even for relatively large graphs; the final MAE on $\theta$ parameters is $0.003$.

\subsection{Beyond Assumption \ref{a:inclusion=in-family}}\label{sec: perturbed f}

In this section, we empirically study whether Assumption \ref{a:inclusion=in-family} is restrictive in practical applications. Specifically, we consider different degrees of model mismatch between the system model in \eqref{eq:system-model} and the approximating model in \eqref{eq:approx-model}. Unless otherwise specified, we use the same dataset and experimental setup as described in Appendix~\ref{appendix:dataset description}. Additional details and results are deferred to Appendix~\ref{appendix:additional figures perturbed f}.
\begin{table}
    \centering
    \footnotesize
    \caption{Calibration of $\pta$ under varying levels of misconfiguration for predictive function $f_\psi$. Results are the mean $\pm$ 1 standard deviation assessed over 8 independent runs.}
    \begin{tabular}{ccc}
        \toprule
        Max pert.\ $\Psi$  & MAE on $\theta$ & Max AE on $\theta$ \\
        \midrule
        0    & 0.009 $\pm$ 0.001   & 0.06 $\pm$ 0.01 \\ 
        0.1  & 0.010 $\pm$ 0.001   & 0.07 $\pm$ 0.01\\ 
        0.2  & 0.012 $\pm$ 0.004   & 0.08 $\pm$ 0.02\\ 
        0.5  & 0.028 $\pm$ 0.011   & 0.16 $\pm$ 0.06\\ 
        0.8  & 0.047 $\pm$ 0.009   & 0.28 $\pm$ 0.06\\ 
        \bottomrule
    \end{tabular}
\label{table: perturbed f}
\end{table}
\paragraph{Perturbed $f_{\psi^*}$}
As a first experiment, we train $\pta$ while keeping the parameters of the predictive function $f_\psi$ fixed to a random perturbation of the data-generating model $f^*=f_{\psi^*}$. A perturbed version of $f_\psi^*$ is built by uniformly drawing independent perturbation scalar values $\delta_i \sim \mathcal{U}[-\Psi, \Psi]$, one for each parameter $\psi_i^*$ of $f_{\psi^*}$. Then, each parameter of GNN $f_\psi$ is given as $\psi_i = (1 + \delta_i) \psi_i^*$. Table~\ref{table: perturbed f} shows that the learned latent distribution remains reasonably calibrated.
Finally, Figures \ref{fig: perturbed 01}-\ref{fig: perturbed 08} show the learned parameter vectors $\theta$ for randomly extracted runs and highlight that the maximum AE of Table~\ref{table: perturbed f} is observed only sporadically.

\paragraph{Generic GNN as $f_\psi$} 
In this second experiment, we set $f_\psi$ to be a generic multilayer GNN 
which we jointly train with graph distribution $\pta$.
The model family $\{f_\psi\}$ employed does not include $f^*$, as
$f^*$ uses L-hop adjacency matrices generated from the sampled adjacency matrix $A$, while $f_\psi$ relies on multiple nonlinear 1-hop layers; additional details are reported in Appendix \ref{appendix:additional figures perturbed f}. Upon convergence, models achieved $\lossfunc < 0.19$ using the MSE as loss function $\ell$ in \eqref{eq:T-bayes-rules}; The performance is in line with results in Figure~\ref{fig: MAE point predictions}. At last, note that as the GNN used adds self-loops, the diagonal elements of the adjacency matrix are learned as zero, resulting in a larger MAE on $\theta$ (see Figure \ref{fig: theta model mismatch}). However, this does not impair learning the off-diagonal $\theta_{ij}$ parameters (i.e., for $i \neq j$). Notably, in the worst-performing model, these off-diagonal parameters achieve a MAE of less than $0.03$, effectively calibrating the latent distribution.

\paragraph{Misconfigured $\pta$} 
Finally, we violate Assumption~\ref{a:inclusion=in-family} by fixing $f_\psi = f^*$ and constraining some components of $\theta$ to incorrect values. Specifically, we force parameters $\theta_{i,j}$ for all edges $i,j$ associated with nodes with id 2 and 3 in Figure~\ref{fig:stoc-gpvar-graph} to $0.25$, instead of the correct value of $\theta_{i,j}^*=0.75$ as in $\psa$. 
Results indicate that the free parameters in $\theta$ are learned appropriately. Notably, increased uncertainty is observed for spurious edges linking to nodes in the first node community (see Figure~\ref{fig:stoc-gpvar-graph}). This is expected given that nearly 60\% of the edges in the community were significantly downsampled. Figures \ref{fig: misconfigured pta 1} and \ref{fig: misconfigured pta 2} in Appendix \ref{par: misconfigured P} show the learned parameters from randomly selected runs.

\begin{table}
    \centering
    \footnotesize
    \caption{Calibration and point-prediction performance of models trained by minimizing different loss functions. 
    Losses $\losssample$ follow the approach proposed in this paper. 
    Bold numbers indicate the best-performing models ($p$-value of the Welch’s t-test $< 0.01$).}
    \begin{tabular}{cccc}
        \toprule
        Train loss  & MAE on $\theta$ & MAE on $y$ & MSE on $y$ \\ 
        \midrule
        $\losslit_{1, \ell: \text{ MAE}} $    & $.087 \pm .001$        & $\mathbf{.270 \pm .003}$   & $.180 \pm .003$  \\ 
        $\losslit_{1, \ell: \text{ MSE}} $    & $.087 \pm .001$        & $.293 \pm .001$            & $\mathbf{.161 \pm .002}$ \\
        $\losslit_{2, \ell: \text{ MAE}} $    & $.086 \pm .001$            & $\mathbf{.270 \pm .002}$            & $.176 \pm .002$ \\ 
        $\losslit_{2, \ell: \text{ MSE}} $    & $.085 \pm .001$            & $.295 \pm .001$            & $\mathbf{.161 \pm .002}$ \\ 
        $\loss^{\text{literature}}_{\text{elbo}}$ & $.082 \pm .001$            & $.310 \pm .010$            & $.191 \pm .020$ \\
        $\lossfunc_{\ell: \text{ MSE}}$       & $.025 \pm .001$            & $\mathbf{.271 \pm .003}$   & $\mathbf{.161 \pm .002}$ \\ 
        \midrule
        $\losssample_{\Delta: \text{ CRPS}}$  & $\mathbf{.010 \pm .002}$   & $\mathbf{.269 \pm .001}$   & $\mathbf{.159 \pm .001}$ \\ 
        $\losssample_{\Delta: \text{ MMD}}$   & $\mathbf{.009 \pm .001}$   & $\mathbf{.269 \pm .001}$   & $\mathbf{.159 \pm .001}$ \\ 
        \bottomrule
    \end{tabular}
\label{table: empirical calibration}
\end{table}

\subsection{Comparison of Loss Functions}\label{sec: comparison with literature}
As a final experiment, we empirically demonstrate that the proposed choice of loss functions \eqref{eq:distribution discrepancy loss} is more effective at calibrating the latent graph distribution, while maintaining or sometimes improving point prediction performance compared to other commonly used loss functions. 

\paragraph{Considered loss functions}
Following our approach we consider two distributional losses, based on the MMD $\losssample_{\Delta: \text{ MMD}}$ and the energy distance
\footnote{By following Section~\ref{sec: Practical Implementation}, the energy distance reduces to the well-known Continuous Ranked Probability Score (CRPS) \citep{gneiting2007strictly}.} 
$\losssample_{\Delta: \text{ CRPS}}$.
As point prediction loss, we use $\lossfunc_{\ell: \text{ MSE}}$ defined in \eqref{eq:T-bayes-rules} based on the mean squared error.
Additionally, we consider three families of losses used in the GSL literature. 
The first one, defined as 
\begin{equation}\label{eq: 3EV loss}
\losslit_{1,\ell} = \E_{x, y^*} \E_{A \sim \pta} \left[\ell(f_\psi(x,A), y^*) \right]
\end{equation}
has been employed, e.g., in \citet{franceschi2019learning} and \citet{cini2023sparse}. Note that, differently from $\lossfunc_{\ell:\text{MSE}}$, the expectation over $A$ is taken outside function $\ell$.
The second family, denoted as $\losslit_{2,\ell}$, is inspired by \citet{kazi2022differentiable}. $\losslit_{2,\ell}$ refines $\losslit_{1,\ell}$ focusing its optimization to node-level predictions; further details follow in Appendix~\ref{appendix: comparison with literature}.
For $\losslit_{1,\ell}$ and $\losslit_{2,\ell}$, we use both MAE and MSE as $\ell$. 

Finally, we adapt the loss function used in \cite{elinas2020variational} for the synthetic regression task:
\begin{multline}\label{eq: Elbo loss}
\loss^{\text{literature}}_{\text{elbo}} = -\E_{x, y^*} \E_{A \sim \pta} \left[ \text{log}(P^\psi_{y|x^*,A}(y^*)) \right] \\
+ KL \left[ P^\theta_A(A) || \bar{P}_A(A) \right]
\end{multline}

where $\bar{P}_A(A)$ is a prior distribution and $P^\psi_{y|x^*,A}(y^*)$ is a Gaussian distribution whose mean vector is determined for each node by the GNN output and standard deviation is set as a hyperparameter. We explored different standard deviations and choices of the prior. Details can be found in Appendix~\ref{appendix: comparison with literature}

\paragraph{Results on point prediction}
Table \ref{table: empirical calibration} shows that models trained with $\losslit_{1,\ell}$, $\losslit_{2,\ell}$ and $\lossfunc_{\ell}$ achieve near-optimal
\footnote{Numerical estimates suggest that the ground-truth optimal MAE and MSE achievable by a predictor are approximately $0.267$ and $0.158$, respectively.}
point predictions according to their respective performance metric (MAE or MSE).
Namely, optimizing $\losslit_{1,\ell:\text{MAE}}$ and $\losslit_{2,\ell:\text{MAE}}$ leads to minimal MAE, but not to minimal MSE; similarly, optimizing $\losslit_{1,\ell:\text{MSE}}$ and $\losslit_{2,\ell:\text{MSE}}$ results in minimal MSE. 
Conversely, predictors trained with either $\losssample_{\Delta: \text{MMD}}$ or $\losssample_{\Delta: \text{CRPS}}$ achieve optimal prediction performance for both metrics.
Interestingly, also $\lossfunc_{\ell:\text{MSE}}$ leads to near-optimal predictions in terms of both MAE and MSE. 
We attribute the superiority of $\lossfunc$ over $\losslit_{1,\ell}$ and $\losslit_{2,\ell}$ to the use of functional $T$ in \eqref{eq:single point prediction using T definition} which enables a more accurate probabilistic modeling of the data-generating process. A similar observation holds for the calibration error, discussed in the next paragraph.

\paragraph{Results on calibration} 
Optimizing the proposed losses ($\losssample_{\Delta: \text{MMD}}$ or $\losssample_{\Delta: \text{CRPS}}$) yields the smallest calibration errors, as measured by the MAE of the latent distribution parameters $\theta$ in $\pta$. In contrast, loss functions commonly used in the literature result in statistically worse calibration performance. Notably, while the point-prediction loss $\lossfunc_{\ell:\text{MSE}}$ outperforms $\losslit_1$ and $\losslit_2$ in terms of calibration error, it remains statistically inferior to the proposed distributional loss $\losssample_\Delta$.

We conclude that the proposed approach was the only one capable of effectively solving the joint learning problem.

\section{Conclusions}\label{sec: Conclusions}

Graph structure learning has emerged as a research field focused on learning graph topologies in support of solving downstream predictive tasks.
Assuming stochastic latent graph structures, we are led to a joint optimization objective: 
(i) accurately learning the distribution of the latent topology while (ii) achieving optimal prediction performance on the downstream task. In this paper, at first, we prove both positive and negative theoretical results to demonstrate that appropriate loss functions must be chosen to solve this joint learning problem.
Second, we propose a sampling-based learning method that does not require the computation of the predictive likelihood. Our empirical results demonstrate that this approach achieves optimal point predictions on the considered downstream task while also yielding calibrated latent graph distributions.

\section*{Impact Statement}
This paper presents work whose goal is to advance the field of Machine Learning. There are many potential societal consequences of our work, none which we feel must be specifically highlighted here.

\section*{Acknowledgments}
This work was supported by the Swiss National Science Foundation project FNS 204061: \emph{HORD GNN: Higher-Order Relations and Dynamics in Graph Neural Networks} and partly supported by International Partnership Program of the Chinese Academy of Sciences under Grant
104GJHZ2022013GC.

\bibliographystyle{icml2025}

\begin{thebibliography}{63}
\providecommand{\natexlab}[1]{#1}
\providecommand{\url}[1]{\texttt{#1}}
\expandafter\ifx\csname urlstyle\endcsname\relax
  \providecommand{\doi}[1]{doi: #1}\else
  \providecommand{\doi}{doi: \begingroup \urlstyle{rm}\Url}\fi

\bibitem[Bartler et~al.(2019)Bartler, Wiewel, Mauch, and Yang]{bartler2019training}
Bartler, A., Wiewel, F., Mauch, L., and Yang, B.
\newblock Training variational autoencoders with discrete latent variables using importance sampling.
\newblock In \emph{2019 27th European Signal Processing Conference (EUSIPCO)}, pp.\  1--5. IEEE, 2019.

\bibitem[Berger(1990)]{berger1990statistical}
Berger, J.~O.
\newblock Statistical decision theory.
\newblock In \emph{Time Series and Statistics}, pp.\  277--284. Springer, 1990.

\bibitem[Bi{\'n}kowski et~al.(2018)Bi{\'n}kowski, Sutherland, Arbel, and Gretton]{binkowski2018demystifying}
Bi{\'n}kowski, M., Sutherland, D.~J., Arbel, M., and Gretton, A.
\newblock Demystifying mmd gans.
\newblock In \emph{International Conference on Learning Representations}, 2018.

\bibitem[Cini et~al.(2023)Cini, Zambon, and Alippi]{cini2023sparse}
Cini, A., Zambon, D., and Alippi, C.
\newblock Sparse graph learning from spatiotemporal time series.
\newblock \emph{Journal of Machine Learning Research}, 24:\penalty0 1--36, 2023.

\bibitem[Connor et~al.(2021)Connor, Canal, and Rozell]{connor2021variational}
Connor, M., Canal, G., and Rozell, C.
\newblock Variational autoencoder with learned latent structure.
\newblock In \emph{International Conference on Artificial Intelligence and Statistics}, pp.\  2359--2367. PMLR, 2021.

\bibitem[Coutino et~al.(2020)Coutino, Isufi, Maehara, and Leus]{coutino2020state}
Coutino, M., Isufi, E., Maehara, T., and Leus, G.
\newblock State-space network topology identification from partial observations.
\newblock \emph{IEEE Transactions on Signal and Information Processing over Networks}, 6:\penalty0 211--225, 2020.

\bibitem[De~Felice et~al.(2024)De~Felice, Cini, Zambon, Gusev, and Alippi]{defelice2024graphbased}
De~Felice, G., Cini, A., Zambon, D., Gusev, V., and Alippi, C.
\newblock Graph-based {{Virtual Sensing}} from {{Sparse}} and {{Partial Multivariate Observations}}.
\newblock In \emph{The {{Twelfth International Conference}} on {{Learning Representations}}}, 2024.

\bibitem[Deleu et~al.(2022)Deleu, G{\'o}is, Emezue, Rankawat, Lacoste-Julien, Bauer, and Bengio]{deleu2022bayesian}
Deleu, T., G{\'o}is, A., Emezue, C., Rankawat, M., Lacoste-Julien, S., Bauer, S., and Bengio, Y.
\newblock Bayesian structure learning with generative flow networks.
\newblock In \emph{Uncertainty in Artificial Intelligence}, pp.\  518--528. PMLR, 2022.

\bibitem[Dong et~al.(2016)Dong, Thanou, Frossard, and Vandergheynst]{dong2016learning}
Dong, X., Thanou, D., Frossard, P., and Vandergheynst, P.
\newblock Learning laplacian matrix in smooth graph signal representations.
\newblock \emph{IEEE Transactions on Signal Processing}, 64\penalty0 (23):\penalty0 6160--6173, 2016.

\bibitem[Elinas et~al.(2020)Elinas, Bonilla, and Tiao]{elinas2020variational}
Elinas, P., Bonilla, E.~V., and Tiao, L.
\newblock Variational inference for graph convolutional networks in the absence of graph data and adversarial settings.
\newblock \emph{Advances in Neural Information Processing Systems}, 33:\penalty0 18648--18660, 2020.

\bibitem[Fatemi et~al.(2021)Fatemi, El~Asri, and Kazemi]{fatemi2021slaps}
Fatemi, B., El~Asri, L., and Kazemi, S.~M.
\newblock Slaps: Self-supervision improves structure learning for graph neural networks.
\newblock \emph{Advances in Neural Information Processing Systems}, 34:\penalty0 22667--22681, 2021.

\bibitem[Fey \& Lenssen(2019)Fey and Lenssen]{fey2019fast}
Fey, M. and Lenssen, J.~E.
\newblock Fast graph representation learning with pytorch geometric.
\newblock \emph{arXiv preprint arXiv:1903.02428}, 2019.

\bibitem[Fout et~al.(2017)Fout, Byrd, Shariat, and Ben-Hur]{fout2017protein}
Fout, A., Byrd, J., Shariat, B., and Ben-Hur, A.
\newblock Protein interface prediction using graph convolutional networks.
\newblock \emph{Advances in neural information processing systems}, 30, 2017.

\bibitem[Franceschi et~al.(2019)Franceschi, Niepert, Pontil, and He]{franceschi2019learning}
Franceschi, L., Niepert, M., Pontil, M., and He, X.
\newblock Learning discrete structures for graph neural networks.
\newblock In \emph{International conference on machine learning}, pp.\  1972--1982. PMLR, 2019.

\bibitem[Gneiting(2011)]{gneiting2011making}
Gneiting, T.
\newblock Making and {{Evaluating Point Forecasts}}.
\newblock \emph{Journal of the American Statistical Association}, 106\penalty0 (494):\penalty0 746--762, June 2011.
\newblock ISSN 0162-1459.
\newblock \doi{10.1198/jasa.2011.r10138}.

\bibitem[Gneiting \& Raftery(2007)Gneiting and Raftery]{gneiting2007strictly}
Gneiting, T. and Raftery, A.~E.
\newblock Strictly {{Proper Scoring Rules}}, {{Prediction}}, and {{Estimation}}.
\newblock \emph{Journal of the American Statistical Association}, 102\penalty0 (477):\penalty0 359--378, 2007.
\newblock ISSN 0162-1459.

\bibitem[Gomez~Rodriguez et~al.(2013)Gomez~Rodriguez, Leskovec, and Sch{\"o}lkopf]{gomez2013structure}
Gomez~Rodriguez, M., Leskovec, J., and Sch{\"o}lkopf, B.
\newblock Structure and dynamics of information pathways in online media.
\newblock In \emph{Proceedings of the sixth ACM international conference on Web search and data mining}, pp.\  23--32, 2013.

\bibitem[Gray et~al.(2020)Gray, Mitchell, and Roughan]{gray2020bayesian}
Gray, C., Mitchell, L., and Roughan, M.
\newblock Bayesian inference of network structure from information cascades.
\newblock \emph{IEEE Transactions on Signal and Information Processing over Networks}, 6:\penalty0 371--381, 2020.

\bibitem[Gretton et~al.(2012)Gretton, Borgwardt, Rasch, Sch{\"o}lkopf, and Smola]{gretton2012kernel}
Gretton, A., Borgwardt, K.~M., Rasch, M.~J., Sch{\"o}lkopf, B., and Smola, A.
\newblock A kernel two-sample test.
\newblock \emph{The Journal of Machine Learning Research}, 13\penalty0 (1):\penalty0 723--773, 2012.

\bibitem[Guo et~al.(2017)Guo, Pleiss, Sun, and Weinberger]{guo2017calibration}
Guo, C., Pleiss, G., Sun, Y., and Weinberger, K.~Q.
\newblock On calibration of modern neural networks.
\newblock In \emph{International conference on machine learning}, pp.\  1321--1330. PMLR, 2017.

\bibitem[Hadjeres et~al.(2017)Hadjeres, Nielsen, and Pachet]{hadjeres2017glsr}
Hadjeres, G., Nielsen, F., and Pachet, F.
\newblock Glsr-vae: Geodesic latent space regularization for variational autoencoder architectures.
\newblock In \emph{2017 IEEE symposium series on computational intelligence (SSCI)}, pp.\  1--7. IEEE, 2017.

\bibitem[Harris et~al.(2020)Harris, Millman, Van Der~Walt, Gommers, Virtanen, Cournapeau, Wieser, Taylor, Berg, Smith, et~al.]{harris2020array}
Harris, C.~R., Millman, K.~J., Van Der~Walt, S.~J., Gommers, R., Virtanen, P., Cournapeau, D., Wieser, E., Taylor, J., Berg, S., Smith, N.~J., et~al.
\newblock Array programming with numpy.
\newblock \emph{Nature}, 585\penalty0 (7825):\penalty0 357--362, 2020.

\bibitem[Hunter(2007)]{hunter2007matplotlib}
Hunter, J.~D.
\newblock Matplotlib: A 2d graphics environment.
\newblock \emph{Computing in science \& engineering}, 9\penalty0 (03):\penalty0 90--95, 2007.

\bibitem[Jiang et~al.(2019)Jiang, Zhang, Lin, Tang, and Luo]{jiang2019semi}
Jiang, B., Zhang, Z., Lin, D., Tang, J., and Luo, B.
\newblock Semi-supervised learning with graph learning-convolutional networks.
\newblock In \emph{Proceedings of the IEEE/CVF conference on computer vision and pattern recognition}, pp.\  11313--11320, 2019.

\bibitem[Joo et~al.(2020)Joo, Lee, Park, and Moon]{joo2020dirichlet}
Joo, W., Lee, W., Park, S., and Moon, I.-C.
\newblock Dirichlet variational autoencoder.
\newblock \emph{Pattern Recognition}, 107:\penalty0 107514, 2020.

\bibitem[Kalofolias(2016)]{kalofolias2016learn}
Kalofolias, V.
\newblock How to learn a graph from smooth signals.
\newblock In \emph{Artificial intelligence and statistics}, pp.\  920--929. PMLR, 2016.

\bibitem[Kazi et~al.(2022)Kazi, Cosmo, Ahmadi, Navab, and Bronstein]{kazi2022differentiable}
Kazi, A., Cosmo, L., Ahmadi, S.-A., Navab, N., and Bronstein, M.~M.
\newblock Differentiable graph module (dgm) for graph convolutional networks.
\newblock \emph{IEEE Transactions on Pattern Analysis and Machine Intelligence}, 45\penalty0 (2):\penalty0 1606--1617, 2022.

\bibitem[Kingma \& Ba(2014)Kingma and Ba]{kingma2014adam}
Kingma, D.~P. and Ba, J.
\newblock Adam: A method for stochastic optimization.
\newblock \emph{arXiv preprint arXiv:1412.6980}, 2014.

\bibitem[Kingma \& Welling(2013)Kingma and Welling]{kingma2013auto}
Kingma, D.~P. and Welling, M.
\newblock Auto-encoding variational bayes.
\newblock \emph{arXiv preprint arXiv:1312.6114}, 2013.

\bibitem[Kipf et~al.(2018)Kipf, Fetaya, Wang, Welling, and Zemel]{kipf2018neural}
Kipf, T., Fetaya, E., Wang, K.-C., Welling, M., and Zemel, R.
\newblock Neural relational inference for interacting systems.
\newblock In \emph{International conference on machine learning}, pp.\  2688--2697. PMLR, 2018.

\bibitem[Kuleshov et~al.(2018)Kuleshov, Fenner, and Ermon]{kuleshov2018accurate}
Kuleshov, V., Fenner, N., and Ermon, S.
\newblock Accurate uncertainties for deep learning using calibrated regression.
\newblock In \emph{International conference on machine learning}, pp.\  2796--2804. PMLR, 2018.

\bibitem[Laves et~al.(2020)Laves, Ihler, Fast, Kahrs, and Ortmaier]{laves2020well}
Laves, M.-H., Ihler, S., Fast, J.~F., Kahrs, L.~A., and Ortmaier, T.
\newblock Well-calibrated regression uncertainty in medical imaging with deep learning.
\newblock In \emph{Medical imaging with deep learning}, pp.\  393--412. PMLR, 2020.

\bibitem[Lokhov(2016)]{lokhov2016reconstructing}
Lokhov, A.
\newblock Reconstructing parameters of spreading models from partial observations.
\newblock \emph{Advances in Neural Information Processing Systems}, 29, 2016.

\bibitem[Mateos et~al.(2019)Mateos, Segarra, Marques, and Ribeiro]{mateos2019connecting}
Mateos, G., Segarra, S., Marques, A.~G., and Ribeiro, A.
\newblock Connecting the dots: Identifying network structure via graph signal processing.
\newblock \emph{IEEE Signal Processing Magazine}, 36\penalty0 (3):\penalty0 16--43, 2019.

\bibitem[Mnih et~al.(2016)Mnih, Badia, Mirza, Graves, Lillicrap, Harley, Silver, and Kavukcuoglu]{mnih2016asynchronous}
Mnih, V., Badia, A.~P., Mirza, M., Graves, A., Lillicrap, T., Harley, T., Silver, D., and Kavukcuoglu, K.
\newblock Asynchronous methods for deep reinforcement learning.
\newblock In \emph{International conference on machine learning}, pp.\  1928--1937. PMLR, 2016.

\bibitem[Mohamed \& Lakshminarayanan(2016)Mohamed and Lakshminarayanan]{mohamed2016learning}
Mohamed, S. and Lakshminarayanan, B.
\newblock Learning in implicit generative models.
\newblock \emph{arXiv preprint arXiv:1610.03483}, 2016.

\bibitem[Mohamed et~al.(2020)Mohamed, Rosca, Figurnov, and Mnih]{mohamed2020monte}
Mohamed, S., Rosca, M., Figurnov, M., and Mnih, A.
\newblock Monte carlo gradient estimation in machine learning.
\newblock \emph{The Journal of Machine Learning Research}, 21\penalty0 (1):\penalty0 5183--5244, 2020.

\bibitem[Morris et~al.(2019)Morris, Ritzert, Fey, Hamilton, Lenssen, Rattan, and Grohe]{morris2019weisfeiler}
Morris, C., Ritzert, M., Fey, M., Hamilton, W.~L., Lenssen, J.~E., Rattan, G., and Grohe, M.
\newblock Weisfeiler and leman go neural: Higher-order graph neural networks.
\newblock In \emph{Proceedings of the AAAI conference on artificial intelligence}, volume~33, pp.\  4602--4609, 2019.

\bibitem[Mukhoti et~al.(2020)Mukhoti, Kulharia, Sanyal, Golodetz, Torr, and Dokania]{mukhoti2020calibrating_focal_loss}
Mukhoti, J., Kulharia, V., Sanyal, A., Golodetz, S., Torr, P., and Dokania, P.
\newblock Calibrating deep neural networks using focal loss.
\newblock \emph{Advances in Neural Information Processing Systems}, 33:\penalty0 15288--15299, 2020.

\bibitem[M{\"u}ller(1997)]{muller1997integral}
M{\"u}ller, A.
\newblock Integral probability metrics and their generating classes of functions.
\newblock \emph{Advances in applied probability}, 29\penalty0 (2):\penalty0 429--443, 1997.

\bibitem[M{\"u}ller et~al.(2019)M{\"u}ller, Kornblith, and Hinton]{muller2019does_lable_smoothing}
M{\"u}ller, R., Kornblith, S., and Hinton, G.~E.
\newblock When does label smoothing help?
\newblock \emph{Advances in neural information processing systems}, 32, 2019.

\bibitem[Niepert et~al.(2021)Niepert, Minervini, and Franceschi]{niepert2021implicit}
Niepert, M., Minervini, P., and Franceschi, L.
\newblock Implicit {{MLE}}: {{Backpropagating Through Discrete Exponential Family Distributions}}.
\newblock In \emph{Advances in {{Neural Information Processing Systems}}}, volume~34, pp.\  14567--14579. Curran Associates, Inc., 2021.

\bibitem[Paszke et~al.(2019)Paszke, Gross, Massa, Lerer, Bradbury, Chanan, Killeen, Lin, Gimelshein, Antiga, et~al.]{paszke2019pytorch}
Paszke, A., Gross, S., Massa, F., Lerer, A., Bradbury, J., Chanan, G., Killeen, T., Lin, Z., Gimelshein, N., Antiga, L., et~al.
\newblock Pytorch: An imperative style, high-performance deep learning library.
\newblock \emph{Advances in neural information processing systems}, 32, 2019.

\bibitem[Pu et~al.(2021)Pu, Cao, Zhang, Dong, and Chen]{pu2021learning}
Pu, X., Cao, T., Zhang, X., Dong, X., and Chen, S.
\newblock Learning to learn graph topologies.
\newblock \emph{Advances in Neural Information Processing Systems}, 34:\penalty0 4249--4262, 2021.

\bibitem[R{\'e}nyi(1961)]{renyi1961measures}
R{\'e}nyi, A.
\newblock On measures of entropy and information.
\newblock In \emph{Proceedings of the fourth Berkeley symposium on mathematical statistics and probability, volume 1: contributions to the theory of statistics}, volume~4, pp.\  547--562. University of California Press, 1961.

\bibitem[Rezende et~al.(2014)Rezende, Mohamed, and Wierstra]{rezende2014stochastic}
Rezende, D.~J., Mohamed, S., and Wierstra, D.
\newblock Stochastic backpropagation and approximate inference in deep generative models.
\newblock In \emph{International conference on machine learning}, pp.\  1278--1286. PMLR, 2014.

\bibitem[Scarselli et~al.(2008)Scarselli, Gori, Tsoi, Hagenbuchner, and Monfardini]{scarselli2008graph}
Scarselli, F., Gori, M., Tsoi, A.~C., Hagenbuchner, M., and Monfardini, G.
\newblock The graph neural network model.
\newblock \emph{IEEE transactions on neural networks}, 20\penalty0 (1):\penalty0 61--80, 2008.

\bibitem[Shafer \& Vovk(2008)Shafer and Vovk]{shafer2008tutorial}
Shafer, G. and Vovk, V.
\newblock A tutorial on conformal prediction.
\newblock \emph{Journal of Machine Learning Research}, 9\penalty0 (3), 2008.

\bibitem[Shang et~al.(2021)Shang, Chen, and Bi]{shang2021discrete}
Shang, C., Chen, J., and Bi, J.
\newblock Discrete graph structure learning for forecasting multiple time series.
\newblock In \emph{International Conference on Learning Representations}, 2021.

\bibitem[Shlomi et~al.(2020)Shlomi, Battaglia, and Vlimant]{shlomi2020graph}
Shlomi, J., Battaglia, P., and Vlimant, J.-R.
\newblock Graph neural networks in particle physics.
\newblock \emph{Machine Learning: Science and Technology}, 2\penalty0 (2):\penalty0 021001, 2020.

\bibitem[Sutton et~al.(1999)Sutton, McAllester, Singh, and Mansour]{sutton1999policy}
Sutton, R.~S., McAllester, D., Singh, S., and Mansour, Y.
\newblock Policy gradient methods for reinforcement learning with function approximation.
\newblock \emph{Advances in neural information processing systems}, 12, 1999.

\bibitem[Sz{\'e}kely \& Rizzo(2013)Sz{\'e}kely and Rizzo]{szekely2013energy}
Sz{\'e}kely, G.~J. and Rizzo, M.~L.
\newblock Energy statistics: A class of statistics based on distances.
\newblock \emph{Journal of statistical planning and inference}, 143\penalty0 (8):\penalty0 1249--1272, 2013.

\bibitem[Van Den~Oord et~al.(2017)Van Den~Oord, Vinyals, et~al.]{van2017neural}
Van Den~Oord, A., Vinyals, O., et~al.
\newblock Neural discrete representation learning.
\newblock \emph{Advances in neural information processing systems}, 30, 2017.

\bibitem[Wasserman \& Mateos(2024)Wasserman and Mateos]{wasserman2024graph}
Wasserman, M. and Mateos, G.
\newblock Graph structure learning with interpretable bayesian neural networks.
\newblock \emph{Transactions on machine learning research}, 2024.

\bibitem[Williams(1992)]{williams1992simple}
Williams, R.~J.
\newblock Simple statistical gradient-following algorithms for connectionist reinforcement learning.
\newblock \emph{Machine learning}, 8:\penalty0 229--256, 1992.

\bibitem[Wu et~al.(2019)Wu, Pan, Long, Jiang, and Zhang]{wu2019graph}
Wu, Z., Pan, S., Long, G., Jiang, J., and Zhang, C.
\newblock Graph wavenet for deep spatial-temporal graph modeling.
\newblock In \emph{Proceedings of the 28th International Joint Conference on Artificial Intelligence}, pp.\  1907--1913, 2019.

\bibitem[Wu et~al.(2020)Wu, Pan, Long, Jiang, Chang, and Zhang]{wu2020connecting}
Wu, Z., Pan, S., Long, G., Jiang, J., Chang, X., and Zhang, C.
\newblock Connecting the dots: Multivariate time series forecasting with graph neural networks.
\newblock In \emph{Proceedings of the 26th ACM SIGKDD international conference on knowledge discovery \& data mining}, pp.\  753--763, 2020.

\bibitem[Xu \& Durrett(2018)Xu and Durrett]{xu2018spherical}
Xu, J. and Durrett, G.
\newblock Spherical latent spaces for stable variational autoencoders.
\newblock \emph{arXiv preprint arXiv:1808.10805}, 2018.

\bibitem[Yu et~al.(2021)Yu, Zhang, Jiang, Wu, and Yang]{yu2021graph}
Yu, D., Zhang, R., Jiang, Z., Wu, Y., and Yang, Y.
\newblock Graph-revised convolutional network.
\newblock In \emph{Machine Learning and Knowledge Discovery in Databases: European Conference, ECML PKDD 2020, Ghent, Belgium, September 14--18, 2020, Proceedings, Part III}, pp.\  378--393. Springer, 2021.

\bibitem[Zadrozny \& Elkan(2001)Zadrozny and Elkan]{zadrozny2001obtaining}
Zadrozny, B. and Elkan, C.
\newblock Obtaining calibrated probability estimates from decision trees and naive bayesian classifiers.
\newblock In \emph{Proceedings of the Eighteenth International Conference on Machine Learning}, pp.\  609--616, 2001.

\bibitem[Zhang et~al.(2019)Zhang, Pal, Coates, and Ustebay]{zhang2019bayesian}
Zhang, Y., Pal, S., Coates, M., and Ustebay, D.
\newblock Bayesian graph convolutional neural networks for semi-supervised classification.
\newblock In \emph{Proceedings of the AAAI conference on artificial intelligence}, volume~33, pp.\  5829--5836, 2019.

\bibitem[Zheng et~al.(2013)Zheng, Liu, and Hsieh]{zheng2013u}
Zheng, Y., Liu, F., and Hsieh, H.-P.
\newblock U-air: When urban air quality inference meets big data.
\newblock In \emph{Proceedings of the 19th ACM SIGKDD international conference on Knowledge discovery and data mining}, pp.\  1436--1444, 2013.

\bibitem[Zhu et~al.(2021)Zhu, Xu, Zhang, Liu, Wu, and Wang]{zhu2021deep}
Zhu, Y., Xu, W., Zhang, J., Liu, Q., Wu, S., and Wang, L.
\newblock Deep graph structure learning for robust representations: A survey.
\newblock \emph{arXiv preprint arXiv:2103.03036}, 14:\penalty0 1--1, 2021.

\end{thebibliography}

\newpage
\appendix
\onecolumn

\section{Proofs of the Theoretical Results}\label{appendix:proofs}

\subsection{Minimizing $\loss^{point}$ does not guarantee calibration}\label{appendix: optimal prediction don't guarantee calibration}
In this section, we prove Proposition \ref{th:optimal prediction don't guarantee calibration}.
\begin{proof}[Proof of Proposition \ref{th:optimal prediction don't guarantee calibration}]
Recall the definition of $\loss^{point}$ in \eqref{eq:T-bayes-rules} using \eqref{eq:single point prediction using T definition}
    \[
    \loss^{point}(\psi,\theta)=\E_x \Big[\E_{y^* \sim P^*_{y|x}} \Big[ \ell\big(y^*, T\big[\ptpy\big] \big) \Big] \Big]
    \]
Given loss function $\ell$, $T$ is, by definition \citep{berger1990statistical, gneiting2011making}, the functional that minimizes
\[
\E_{y^* \sim P^*_{y|x}} \Big[ \ell\big(y^*, T\big[\psy\big] \big) \Big]
\]
Therefore, if $\ptpy = \psy \implies \loss^{point}$ is minimal. If another distribution over $y$, namely, $P^{\psi',\theta'}_{y|x}$ parametrized by $\theta'$ and $\psi'$ satisfies:
\[
T\Big[ P^{\psi',\theta'}_{y|x}\Big] = T\Big[\psy\Big]
\]
almost surely on $x$, then,
\begin{align*}
    \nonumber
    \loss^{point}(\theta', \psi') &= \E_x \Big[\E_{y^* \sim P^*_{y|x}} \Big[ \ell\big(y^*, T\big[P^{\psi',\theta'}_{y|x}\big] \big) \Big] \Big] \\ \nonumber
    &= \E_x \Big[\E_{y^* \sim P^*_{y|x}} \Big[ \ell\big(y^*, T\big[\psy\big] \big) \Big] \Big] 
\end{align*}
Thus, $P^{\psi',\theta'}_{y|x}$ minimizes $\loss^{point}$.

Appendix~\ref{appendix MAE and MSE no calibration} discusses graph distributions as counterexamples where $T\big[ P^{\psi',\theta'}_{y|x}\big] = T\big[\psy\big]$ but $P^{\psi',\theta'}_{y|x} \not= \psy$. By this, we conclude that reaching the minimum of $\loss^{point}(\psi,\theta)$ does not always imply $P^{\psi,\theta}_{y|x}=P^*_{y|x}$.
\end{proof}

\subsection{Minimizing $\loss^{point}$ does not guarantee calibration: an example with MAE}\label{appendix MAE and MSE no calibration}

This section shows that $\loss^{point}$ equipped with MAE as $\ell$ admits multiple global minima for different parameters $\theta$, even for simple models and $f_\psi=f^*$.

Consider a single Bernoulli of parameter $\theta^*>1/2$ as latent variable $A$ and a scalar function $f^*(x,A)$ such that $f^*(x,1)>f^*(x,0)$ for all $x$. 
Given input $x$ the value of functional $T(\psy)$ that minimizes 
$$\E_{y\sim\psy} \Big[ \left| y-T\big[\psy\big] \right| \Big] =\theta^* \left| f^*(x,1)-T\big[\psy\big] \right| +(1-\theta^*)\left|f^*(x,0)-T\big[\psy\big] \right|$$
is $T(\psy)=f^*(x,1)$; this derives from the fact that range of $f^*$ is $\{f^*(x,0),f^*(x,1)\}$ and the likelihood of $f^*(x,1)$ is larger than that of $f^*(x,0)$.

Note that $T\big[\psy\big]=f^*(x,1)$ for all $x$, therefore also $\loss^{point}$ is minimized by such $T$. Moreover, $T\big[\psy\big]$ is function of $\theta^*$ and equal to $f^*(x,1)$ for all $\theta>1/2$. We conclude that for any $\theta\ne \theta^*$ distributions $\ptpy$ and $\psy$ are different, yet both of them minimize $\loss^{point}$ if $\theta >1/2$.

A similar reasoning applies for $\theta^* < 1/2$.

\subsection{Minimizing $\loss^{dist}$ guarantees calibration and optimal point predictions}\label{appendix: l dist for calibration pt 1}

This section proves Theorem~\ref{theo:Ldist-Lpoint-Lcal} and a corollary of it.

\begin{proof}[Proof of Theorem~\ref{theo:Ldist-Lpoint-Lcal}]
Recall from Equation \eqref{eq:distribution discrepancy loss} that
\[
\loss^{dist}(\theta) = \E_x\Big[ \Delta(P_{y|x}^*,P_{y|x}^{\theta,\psi}) \Big]
\]

We start by proving that if $\loss^{dist}(\theta,\psi) = 0 \implies \loss^{point}(\theta,\psi) \text{ is minimal}$.

Note that $\loss^{dist}(\theta, \psi) = 0$ implies that $\Delta(P_{y|x}^*,P_{y|x}^{\theta,\psi}) = 0$ almost surely in $x$. Then, by Assumption \ref{a:Delta-premetric}, $P_{y|x}^*=P_{y|x}^{\theta,\psi}$ almost surely on $x$ and, in particular, $T[P_{y|x}^*]=T[P_{y|x}^{\theta,\psi}]$, which leads to $\loss^{point}(\psi,\theta)$ being minimal (Proposition \ref{th:optimal prediction don't guarantee calibration}). 

We now prove that if $\loss^{dist}(\theta,\psi^*) = 0 \implies \loss^{cal}(\theta) = 0$.

From the previous step, we have that $\loss^{dist}(\theta, \psi) = 0$ implies $P_{y|x}^*=P_{y|x}^{\theta,\psi}$ almost surely for $x\in I$. Under the assumption that $f_\psi=f_*$ and the injectivity of $f_*$ in such $x\in I$, for any output $y$ a single $A$ exists such that $f_*(x,A)=y$. Therefore, the probability mass function of $y$ equals that of $A$. Accordingly, $P_{y|x}^*=P_{y|x}^{\theta,\psi}$ implies $P_{A}^*=P_{A}^{\theta}$.

\end{proof}

\label{appendix: l dist for calibration pt 2}

We also prove a corollary of Theorem \ref{theo:Ldist-Lpoint-Lcal}.

\begin{corollary}
\label{corollary: l dist for calibration pt 2} 
Under Assumptions~\ref{a:inclusion=in-family} and \ref{a:Delta-premetric}, 
if 
\begin{enumerate}[itemsep=0.5pt,topsep=0pt]
    \item $\exists \bar x\in \textit{Supp}(P^*_{x})\subseteq \mathcal X$ such that $f^*({\bar x};\cdot)$ is injective,
    \item $f^*(x,A)$ is continuous in $ x$, $\forall A\in \mathcal{A}$,
\end{enumerate} 
then 
\begin{align*}
\mathcal L^{dist}(\theta,\psi^*) = 0 &\implies
\begin{cases}\mathcal L^{point}(\theta,\psi^*) \text{ is minimal}
\\
\mathcal L^{cal}(\theta) =0.
\end{cases}
\end{align*}
\end{corollary}
The corollary shows that it is sufficient that $f^*$ is continuous in $x$ and there exists one point $\bar x$ where $f^*(\bar x,{}\cdot{})$ is injective to meet theorem's hypothesis $\mathbb P_{x\sim P_x^*}(I)>0$; we observe that, as $\mathcal A$ is discrete, the injectivity assumption is not as restrictive as if the domain were continuous.

\begin{proof}
As $\mathcal A$ is a finite set, the minimum $\bar\epsilon=\min_{A\ne A'\in\mathcal A} \lVert f^*(\bar x,A)-f^*(\bar x,A')\rVert>0$ exists and, by the injectivity assumption, is strictly positive.

By continuity of $f^*({}\cdot{},A)$, for every $\epsilon<\frac{1}{2}\bar\epsilon$ there exists $\delta$, such that for all $x\in B(\bar x, \delta)$ we have $\lVert f^*(\bar x,A)-f^*(x,A)\rVert <\epsilon$. It follows that, $\forall x\in B$,
\begin{align*}
\lVert f^*(x,A)&-f^*(x,A')\rVert \\&\ge 
    \lVert f^*({\bar x},A)-f^*({\bar x},A')\rVert 
    - \lVert f^*({\bar x},A)-f^*(x,A)\rVert 
    - \lVert f^*(\bar x,A')-f^*(x,A')\rVert
\\ &\ge
    \lVert f^*({\bar x},A)-f^*({\bar x},A')\rVert
    - 2\epsilon
\\ &\ge
    \lVert f^*({\bar x},A)-f^*({\bar x},A')\rVert
    - \bar\epsilon >0.
\end{align*}

Where the second inequality holds for the continuity of $x \mapsto f^*(x, A)$ $\forall A$. Finally, as $\bar x\in {\rm Supp}(P_x^*)$ and $B(\bar x, \delta)\subseteq I$, we conclude that 
\[
\mathbb P_x(I) \ge \mathbb P_x(B(\bar x, \delta))> 0,
\]
therefore, we are in the hypothesis of Theorem \ref{theo:Ldist-Lpoint-Lcal} and can conclude that
\begin{align*}
\mathcal L^{dist}(\theta,\psi^*) = 0 &\implies
\begin{cases}\mathcal L^{point}(\theta,\psi^*) \text{ is minimal}
\\
\mathcal L^{cal}(\theta) =0.
\end{cases}
\end{align*}
\end{proof}

\subsection{Injectivity hypothesis for graph neural networks}

Now, we show that hypothesis  $\mathbb P_{x\sim P_x^*}(I)>0$ of Theorem~\ref{theo:Ldist-Lpoint-Lcal} is always met for certain families of graph neural networks. 

\begin{proposition}
Consider a 1-layer GNN of the form $f^*(x, A):\sigma(A x) = y$, with $x,y\in\mathbb R^{N}$ and nonlinear bijective activation function $\sigma$. If the support $\text{Supp}(P_x^*)$ of $x$ contains any ball $B$ in $\mathbb R^N$ then $\mathbb P_{x\sim P_x^*}(I)>0$.
    \label{prop: l dist for calibration pt 2}
\end{proposition}

To prove Proposition~\ref{prop: l dist for calibration pt 2}, we rely on following lemma.

\begin{lemma}\label{lemma: x injective iff exists delta}
    Given $g(x, a) = ax$ with $a\in\{0,1\}^{1\times N}$ and $x\in \mathbb R^{N \times 1}$. Let $I_g=\{x:g(x,a) \text{ is injective in } a\}\subseteq \mathcal X$  be the set of points $x\in\mathcal X$ such that map $a \mapsto g(x, a)$ is injective. The following implication holds:
    \begin{equation}
        x \not\in I_g \iff \exists \delta\ne \mathbf{0}  \in \{ -1, 0, 1\}^{1 \times N} \text{ s.t. } \delta \perp x.
    \end{equation}
\end{lemma}

\begin{proof} We prove the two implications separately. 
\begin{description}
\item[$(\implies)$] 
If $x \not\in I_g$, then there exist $a', a'' \in\{0,1\}^{1\times N}$ with $a' \neq a''$ such that  $a'x = a''x$. This implies that $(a' - a'')x = 0 $.
Defining $\delta $ as $ (a' - a'')$, we have proven that there exist $\delta\neq \mathbf 0 \in \{ -1, 0, 1\}^{1 \times N}$ such that $\delta x =0$, i.e., $\delta\perp x$.

\item[$(\impliedby)$]
Assume that $\exists\; \delta \ne \mathbf{0} \in \{ -1, 0, 1\}^{1 \times N}$ such that $\delta \perp x$. Each component $\delta_i$ of $\delta$ can be written as the difference between two values $a'_i, a''_i \in \{0, 1\}$. As $\delta \not = \mathbf{0}$ then there exists at least one index $j\in\{1,\dots,N\}$ such that $a'_j \not= a''_j$. This implies that $\exists\; a', a'' \in\{0,1\}^{1\times N}$ with  $a' \not= a''$ s.t. $(a' - a'')x = 0 $, which implies that $x\not\in I_g$.
\end{description}
\end{proof}

\begin{proof}[Proof of Proposition~\ref{prop: l dist for calibration pt 2}]
We begin by considering the projection $\bar g(x, a) = ax$ with $a\in\{0,1\}^{1\times N}$ and $x\in \R^{N}$. Then we extend to $A\in\{0,1\}^{N\times N}$ and to nonlinear functions.

Let 
$I_{\bar g}^C=\mathbb{R}^N \setminus I_{\bar g}$ be the complement in $\mathbb{R}^N$ of $I_{\bar g}$. 
Recalling Lemma \ref{lemma: x injective iff exists delta} and its notation, we have ${3^N-1}$ possible $\delta$, defining a collection of $(3^N-1)/{2}$ hyperplanes of vectors $x$ perpendicular to at least one $\delta$; set $I_{\bar g}^C$ is the union of such a finite collection of hyperplanes. 
By hypothesis, $\text{Supp}(P_x^*)$ contains a ball $B \in \mathbb{R}^N$, therefore $\text{Supp}(P_x^*) \not \subset I_{\bar g}^C$ and $\mathbb P_{x\sim P_x^*} (I_{\bar g}^C)<1$. We conclude that $\mathbb P_{x\sim P_x^*} (I_{\bar g})=1-\mathbb P_{x\sim P_x^*} (I_{\bar g}^C)>0$.

A similar result is proven for $\bar G(x, A) = Ax$ with $A\in\{0,1\}^{N\times N}$. In fact, $\bar G$ is a stack of $N$ functions $\bar g$ above and $I_{\bar G}=I_{\bar g}$. 
Finally, composing injective function $G$ with injective function $\sigma$ leads to function $g(x, A) = \sigma(G(x, A))$ being injective in $A$ for the same points $x$ for which $G$ is injective, thus proving the proposition.
\end{proof}

\section{Estimation of Optimal $\beta_1$ and $\beta_2$}
\label{appendix:pratical-implementation-optimization}
\label{appendix: optimal beta control variates}
Here we show that, when reducing the variance of the SFE via control variates in \eqref{eq: d loss / d theta REDUCED VARIANCE}, the best $\beta_1$ and $\beta_2$ can be approximated by 
\begin{align}\label{eq:appendix optimal betas}
    \tilde \beta_1 &= \underset{\substack{x\sim P_{x}^*\\ A_1,A_2\sim\pta}}{\E} \Big[ \kappa\left(f_\psi(x ,A_1), f_\psi(x ,A_2) \right) \Big],
    &
    \tilde \beta_2 &= \underset{\substack{(x,y^*)\sim P_{x,y}^*\\ A\sim\pta}}{\E} \Big[ \kappa\left(y^*,f_\psi(x ,A) \right) \Big],
\end{align}

Consider generic function $L(A)$ depending on a sample $A$ of a parametric distribution $\pta(A)$ and the surrogate loss $\tilde L(A)$ in \eqref{eq: control variate equation}, i.e.,
\begin{equation}
    G(A) = L(A)\nabla_\theta \log P^\theta(A) - \beta \Big( h(A) - \E_{A \sim P^\theta} [ h(A)] \Big);
\end{equation}
Following existing literature \cite{sutton1999policy, mnih2016asynchronous} where $\beta$ is often referred to as \emph{baseline} we set $h(A) = \nabla_\theta \log P^\theta(A)$. 
The 1-sample MC approximation of the gradient becomes
\begin{equation}\label{eq: appendix E[F] variance reduced}
\nabla_\theta \E_{A\sim P^\theta}[L(A)] \approx G(A') = ( L(A') - \beta) \nabla_\theta \log P^\theta(A'),
\end{equation}
with $A'$ sampled from $\pta$.
The variance of the estimator is
\begin{multline}
    \V_{A\sim P^\theta} \left[( L(A) - \beta ) \nabla_\theta \log P^\theta(A) \right] = \V_{A\sim P^\theta} \left[ L(A) \nabla_\theta \log P^\theta(A) \right] + 
    \\
    + \beta^2\; \E_{A\sim P^\theta} \left[ \left( \nabla_\theta \log P^\theta(A) \right)^2 \right]
    -2\beta\; \E_{A\sim P^\theta} \left[ L(A) \left( \nabla_\theta \log P^\theta(A) \right)^2 \right]
\end{multline}
and the optimal value $\beta$ that minimizes it is
\begin{equation}
\tilde \beta = \frac{\E_{A\sim P^\theta} \left[ L(A) \left( \nabla_\theta \log P^\theta(A) \right)^2 \right]}{\E_{A\sim P^\theta} \left[ \left( \nabla_\theta \log P^\theta(A) \right)^2 \right]}.
\end{equation}
If we approximate the numerator with $\E[L(A)] \E[(\nabla_\theta \log P^\theta(A) )^2]$, we obtain that $\tilde \beta \approx \E[ L(A)]$. By substituting $L(A)$ with 
the two terms of \eqref{eq: d loss / d theta} we get the values of $\beta_1$ and $\beta_2$ in \eqref{eq:appendix optimal betas}.

We experimentally validate the effectiveness of this choice of $\beta$ in Section \ref{sec: Experiments}.

\section{Further Experimental Details}\label{appendix:experiments}

\subsection{Dataset description and models}\label{appendix:dataset description}

\begin{figure}[!h]
  \centering
  \includegraphics{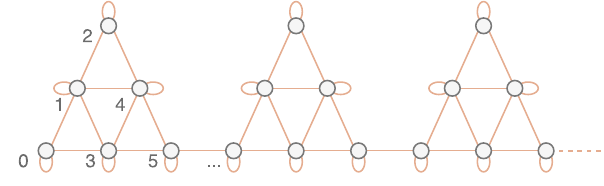}
  \caption{The adjacency matrices used in this paper are sampled from this graph. Each edge in orange is independently sampled with probability $\theta^*$. In the picture, 3 communities of an arbitrarily large graph are shown.}
  \label{fig:community graph}
\end{figure}

\begin{figure}[!h]
  \centering
  \includegraphics[scale=0.35]{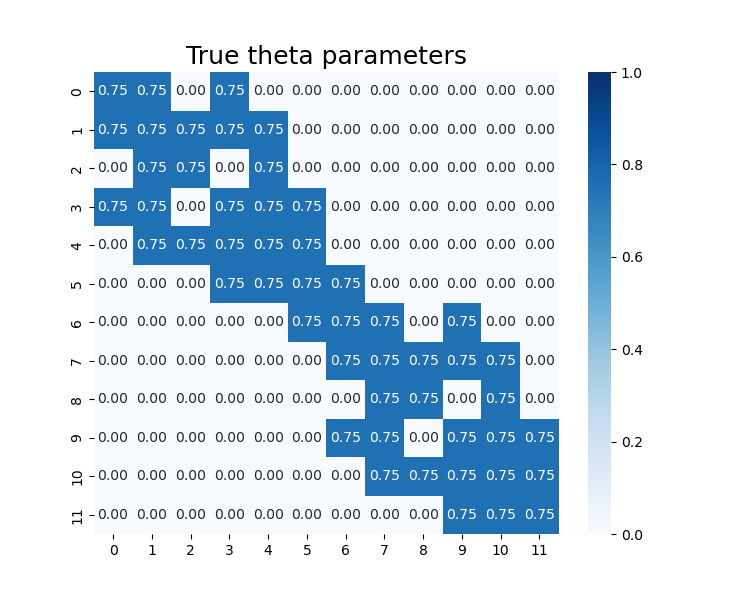}
  \caption{$\theta^*_{ij}$ parameters for each edge of the latent adjacency matrix. Each square corresponds to an edge, and the number inside is the probability of sampling that edge for each prediction.}
  \label{fig:true theta parameters}
\end{figure}

In this section, we describe the considered synthetic dataset, generated from the system model \eqref{eq:system-model}. 
The latent graph distribution $P^*_A$ is a multivariate Bernoulli distribution of parameters $\theta^*_{ij}$: $P^*_A \equiv P_{\theta^*}(A)$ = $\prod_{ij}$ $\theta_{ij}^{* A_{ij}}$ $(1-\theta^*_{ij})^{(1-A_{ij})}$. The components of $\theta^*$ are all null, except for the edges of the graph depicted in Figure~\ref{fig:community graph} which are set to $3/4$.
A heatmap of the adjacency matrix can be found in Figure~\ref{fig:true theta parameters}.

\begin{wraptable}[9]{r}{0.50\textwidth}
\vspace{-7mm}
    \centering
\renewcommand*{\arraystretch}{1.4}
    \centering
    \caption{Table of the parameters used to generate the synthetic dataset.}
    \begin{tabular}{cc}
    \toprule
    Parameter & Values \\
    \midrule
         $\theta^*$ & $0.75$ \\ 
         $\sigma_x$ & $1.5$ \\ 
         $N$ & $12$ \\ 
         $d_{in}$ & $4$ \\ 
         $d_{out}$ & $1$ \\
         $\psi_1^*$ & $[0.3, -0.2, 0.1, -0.2]$ \\ 
         $\psi_2^*$ & $[-0.3, 0.1, 0.2, -0.1]$ \\ \bottomrule
    \end{tabular}
    \vspace{2mm}
    \label{tab:dataset parameters}
\end{wraptable}

Regarding the GNN function $f^*$, we use the following system model:
\begin{equation}
\begin{dcases*}
    y = f_{\psi^*}(A,x) = \text{tanh}\left( \sum_{l=1}^{L} \mathbbm{1}[A^l \centernot= 0] x \psi^*_{l}  \right) \\
    A \sim P_{\theta^*}(A)
\end{dcases*}
\end{equation}
where $\mathbbm{1}[\cdot]$ is the element-wise indicator function: $\mathbbm{1}[a] = 1 \iff a$ is true. 
$x \in \R^{N  \text{ x } d_{in}}$ are randomly generated inputs: $x \sim \mathcal N(0, \sigma_x^2 \mathbb{I})$. 
$\psi^*_{l} \in \R^{d_{out} \text{ x } d_{in}}$ are part of the system model parameters. We summarize the parameters considered in our experiment in Table \ref{tab:dataset parameters}.

The approximating model family \eqref{eq:approx-model} used in the experiment is the same as the data-generating process, with all components of parameter vectors $\theta$ and $\psi$ being trainable.
The squared MMD discrepancy is defined over Rational Quadratic kernel \cite{binkowski2018demystifying} 
\begin{equation}
\kappa(y',y'')=\left(1+\frac{\lVert y' - y''\rVert_2^2}{2\,\alpha\,\sigma^2}\right)^{-\alpha}
\end{equation}
of hyper-parameters $\sigma = 0.04$ and $\alpha = 0.5$ tuned on the validation set.

The model is trained using Adam optimizer \cite{kingma2014adam} with parameters $\beta_1 = 0.9$, $\beta_2 = 0.99$. 
Where not specified, the learning rate is set to $0.05$ and decreased to $0.01$ after 5 epochs. We grouped data points into batches of size 128. Initial values of $\theta$ are independently sampled from the $\mathcal U (0.0, 0.1)$ uniform distribution.

\subsection{Additional details on the experiments of Section \ref{sec:exp:validation}}\label{appendix:additional-figs-validation}
\begin{figure}[h!]
   \begin{minipage}[t]{0.49\textwidth}
     \centering
     \includegraphics[width=\imagewidth\linewidth]{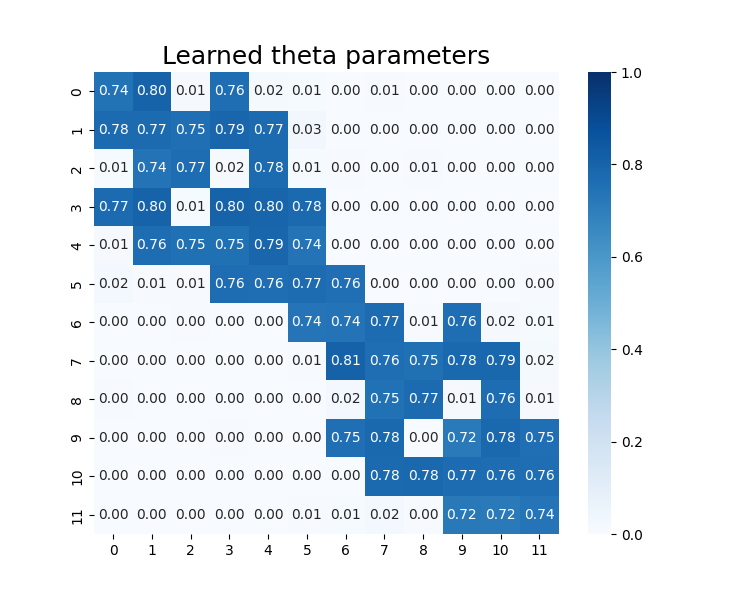}
     \caption{The learned parameters for the latent distribution corresponding to the stochastic adjacency matrix.}\label{fig: theta adj}
   \end{minipage}\hfill
   \begin{minipage}[t]{0.49\textwidth}
     \centering
     \includegraphics[width=\imagewidth\linewidth]{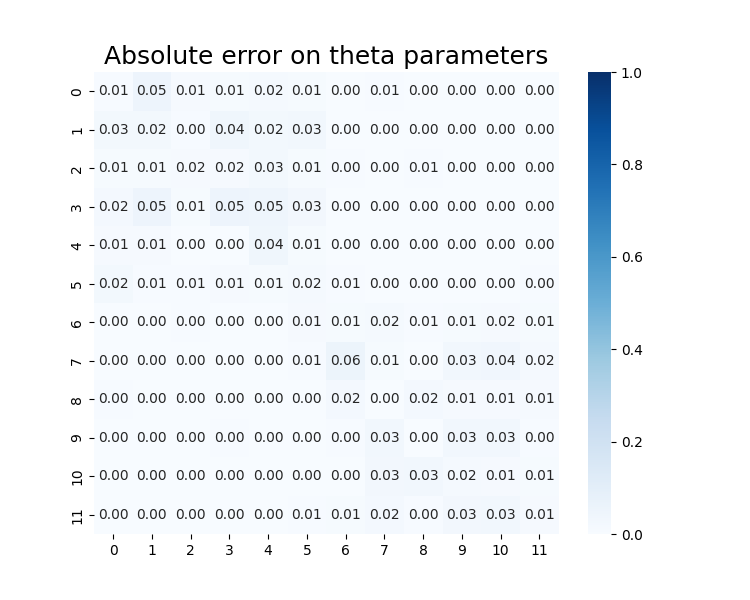}
     \caption{Absolute error made on the parameters of the latent distribution.}\label{fig: ae adj}
   \end{minipage}
\end{figure}

\begin{figure}[!h]
\centering
\includegraphics[width=0.38\textwidth]{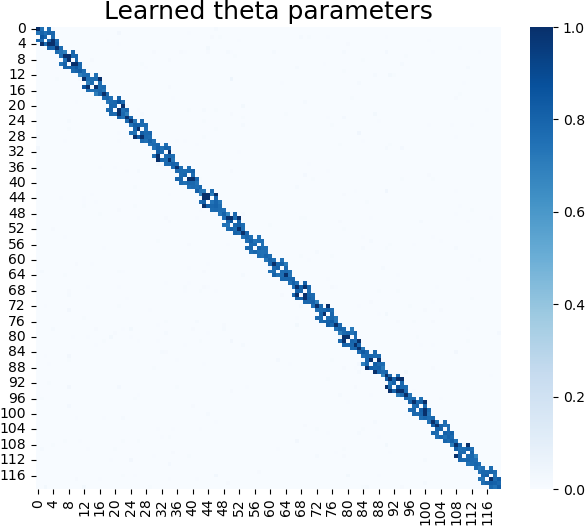}
\caption{Learned $\theta$ parameters for a graph with $\sim 15K$ possible edges.}
\label{fig: larger graph N120}
\end{figure}

We present here additional figures discussed in Section \ref{sec:exp:validation}.
Figure~\ref{fig: theta adj} reports the values of the learned parameters $\theta$, while Figure~\ref{fig: ae adj} the absolute discrepancy from $\theta^*$. 
Figure~\ref{fig: larger graph N120} reports the values of the learned parameters $\theta$ when considering a graph of 120 nodes. 

\subsection{Additional details on the experiments of Section \ref{sec: perturbed f}} \label{appendix:additional figures perturbed f}
We present here additional details discussed in Section \ref{sec: perturbed f}.

\begin{figure}[!h]
\minipage{0.49\textwidth}
\begin{subfigure}[b]{0.49\textwidth}
     \centering
     \includegraphics[width=\linewidth]{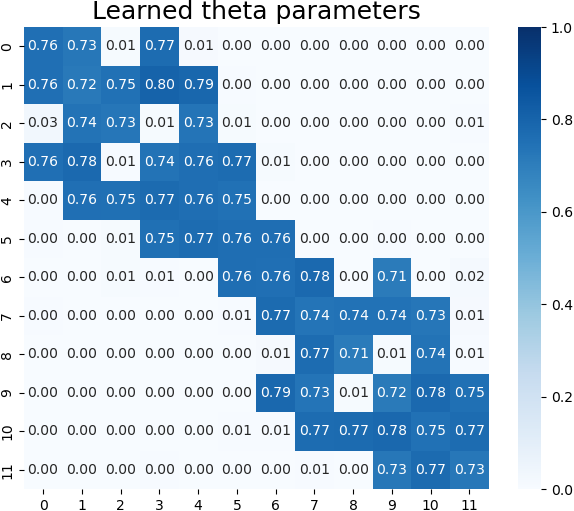}
     \caption{}
\end{subfigure}
\hfill
\begin{subfigure}[b]{0.49\textwidth}
     \centering
     \includegraphics[width=\linewidth]{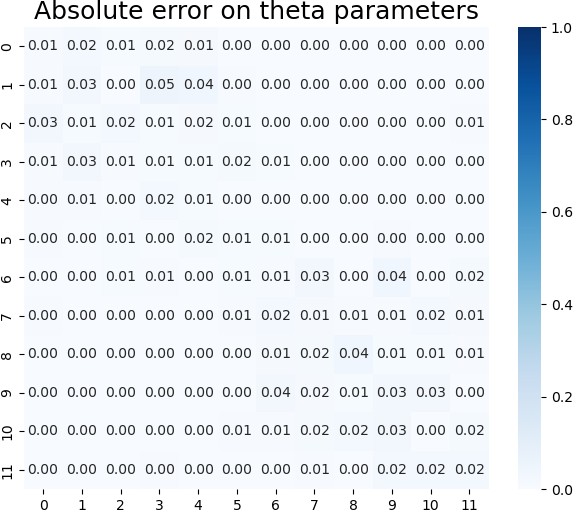}
     \caption{}
\end{subfigure}
\caption{Learned $\theta_{ij}$ parameters (a) and Absolute Error (b) for maximum perturbation factor $\Psi$ of 10\%.}
\label{fig: perturbed 01}
\endminipage\hfill
\minipage{0.49\textwidth}
\begin{subfigure}[b]{0.49\textwidth}
     \centering
     \includegraphics[width=\linewidth]{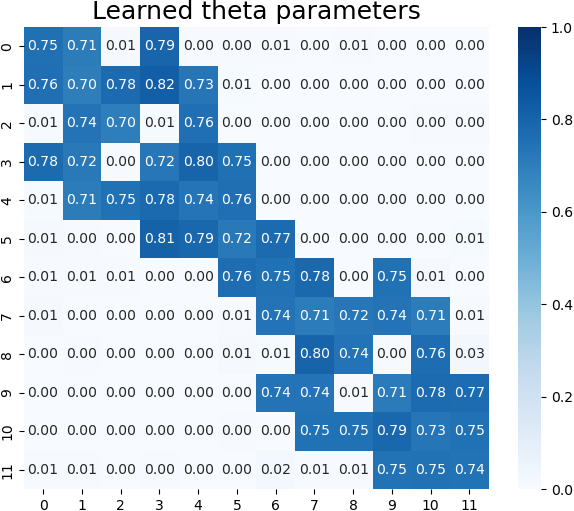}
     \caption{}
\end{subfigure}
\hfill
\begin{subfigure}[b]{0.49\textwidth}
     \centering
     \includegraphics[width=\linewidth]{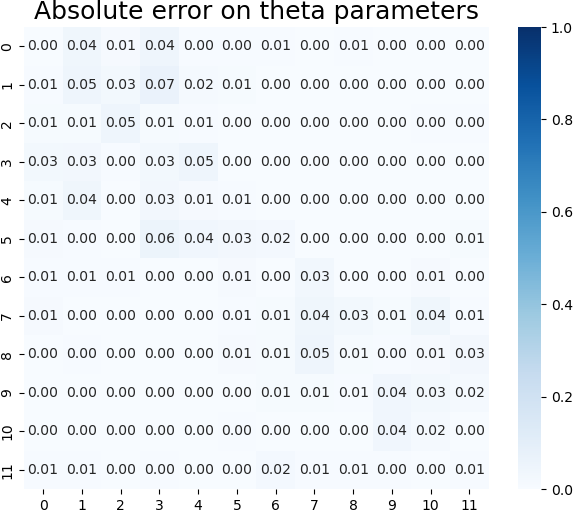}
     \caption{}
\end{subfigure}
\caption{Learned $\theta_{ij}$ parameters (a) and Absolute Error (b) for maximum perturbation factor $\Psi$ of 20\%.}
\label{fig: perturbed 02}
\endminipage\hfill
\end{figure}

\begin{figure}[!h]
\minipage{0.49\textwidth}
\begin{subfigure}[b]{0.49\textwidth}
     \centering
     \includegraphics[width=\linewidth]{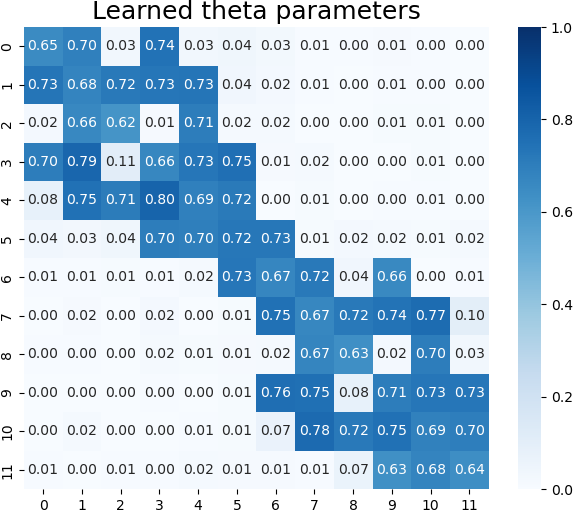}
     \caption{}
\end{subfigure}
\hfill
\begin{subfigure}[b]{0.49\textwidth}
     \centering
     \includegraphics[width=\linewidth]{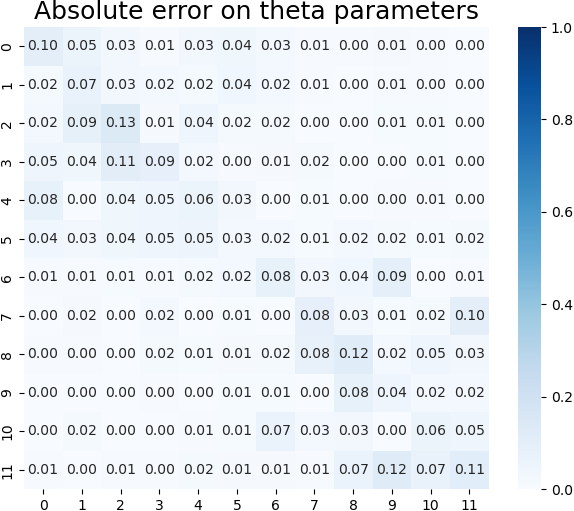}
     \caption{}
\end{subfigure}
\caption{Learned $\theta_{ij}$ parameters (a) and Absolute Error (b) for maximum perturbation factor $\Psi$ of 50\%.}
\label{fig: perturbed 05}
\endminipage\hfill
\minipage{0.49\textwidth}
\begin{subfigure}[b]{0.49\textwidth}
     \centering
     \includegraphics[width=\linewidth]{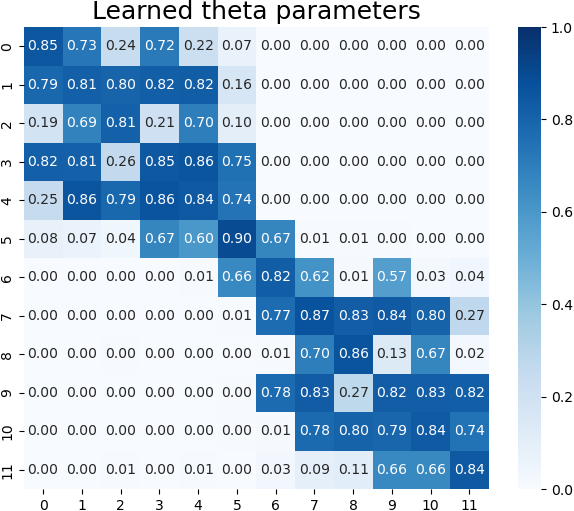}
     \caption{}
\end{subfigure}
\hfill
\begin{subfigure}[b]{0.49\textwidth}
     \centering
     \includegraphics[width=\linewidth]{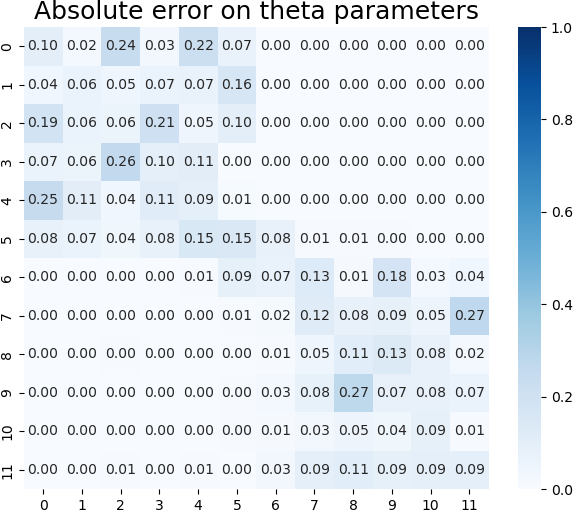}
     \caption{}
\end{subfigure}
\caption{Learned $\theta_{ij}$ parameters (a) and Absolute Error (b) for maximum perturbation factor $\Psi$ of 80\%.}
\label{fig: perturbed 08}
\endminipage\hfill
\end{figure}

\paragraph{Fixed perturbed $f_\psi$} Figures in this paragraph correspond to the experiment where the processing function $f_\psi$ is fixed on a perturbed version of $f^*$. Figures \ref{fig: perturbed 01} $-$ \ref{fig: perturbed 08} correspond to runs with increasing perturbation factor $\Psi$.

\newpage

\begin{wraptable}[12]{r}{0.50\textwidth}
\vspace{-3.7mm}
    \centering
    \footnotesize
    \caption{Network configurations and corresponding convergence results.}
    \begin{tabular}{cc}
        \toprule
        \makecell{Layers dimensions}  & Convergence \\
        \midrule
          $[4, 1]$ & $\mathbf{x}$ \\
          $[4, 1, 1]$ & $\mathbf{x}$ \\
          $[4, 2, 1]$ & \checkmark \\
          $[4, 8, 1]$ & \checkmark \\
          $[4, 8, 2, 1]$ & \checkmark \\
          $[4, 16, 8, 1]$ & \checkmark \\
          $[4, 32, 8, 1]$ & \checkmark \\
          $[4, 64, 8, 1]$ & \checkmark \\
          $[4, 64, 16, 1]$ & \checkmark \\
          $[4, 64, 32, 1]$ & \checkmark \\
          $[4, 8, 8, 4, 1]$ & \checkmark \\
        \bottomrule
    \end{tabular}
\label{table: f_psi GNN}
\end{wraptable}

\paragraph{Generic GNN as $f_\psi$}
To evaluate our approach in a more realistic setting, we use a generic GNN as $f_\psi$. Specifically, we implement GNNs from \cite{morris2019weisfeiler} with varying numbers of layers and layer sizes. It is important to note that the GNN implementation includes self-loops, which prevents the diagonal elements from being correctly learned. However, this does not impede our method from learning the remaining edges accurately.

Table \ref{table: f_psi GNN} presents the network configurations and whether they successfully converged to the ground truth distribution. Since diagonal elements artificially inflate the MAE for $\theta$, we consider a model to have converged if the final MAE on $\theta$ is less than 0.11.

\begin{figure}[!h]
\begin{subfigure}[b]{0.42\textwidth}
     \centering
     \includegraphics[width=0.8\linewidth]{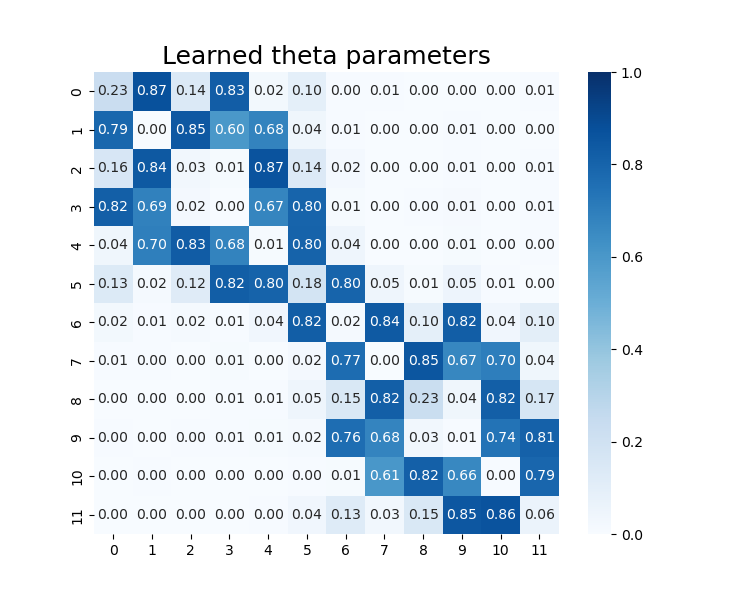}
     \caption{}
\end{subfigure}
\hfill
\begin{subfigure}[b]{0.42\textwidth}
     \centering
     \includegraphics[width=0.8\linewidth]{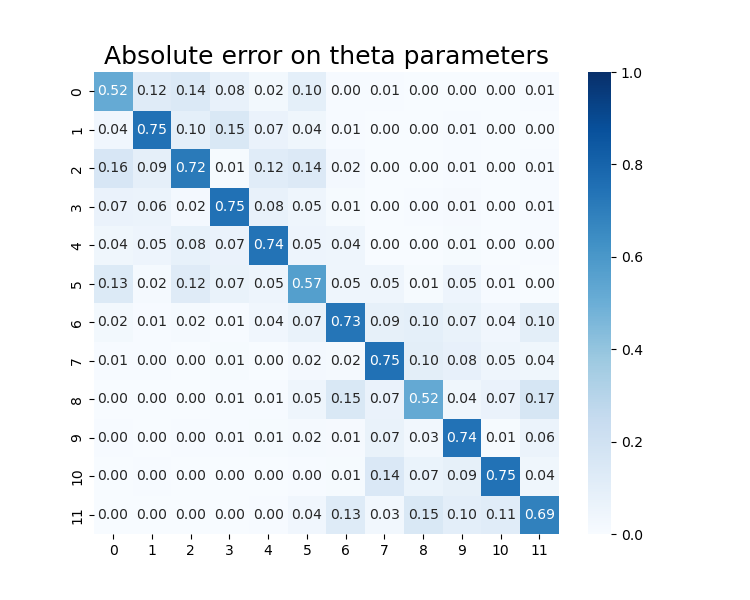}
     \caption{}
\end{subfigure}
\caption{(a) Learned $\theta_{ij}$ parameters when the parametric processing function $f_\psi$ is a generic GNN as presented in \cite{morris2019weisfeiler} and (b) Absolute Error made with respect to true parameters $\theta^*_{ij}$. As self-loops are deterministically added by the network, the diagonal elements should not be considered.}
\label{fig: theta model mismatch}
\end{figure}

Most of the models successfully converged, except those with high bias. This demonstrates that our method is effective even beyond Assumption \ref{a:inclusion=in-family}. In Figure \ref{fig: theta model mismatch} we show the learned parameters of $\pta$ for a randomly extracted run.

\begin{figure}[!h]
\minipage{0.49\textwidth}
\begin{subfigure}[b]{0.49\textwidth}
     \centering
     \includegraphics[width=\linewidth]{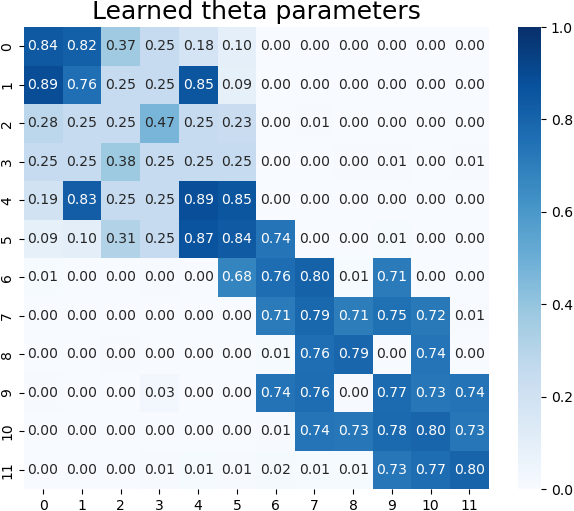}
     \caption{}
\end{subfigure}
\hfill
\begin{subfigure}[b]{0.49\textwidth}
     \centering
     \includegraphics[width=\linewidth]{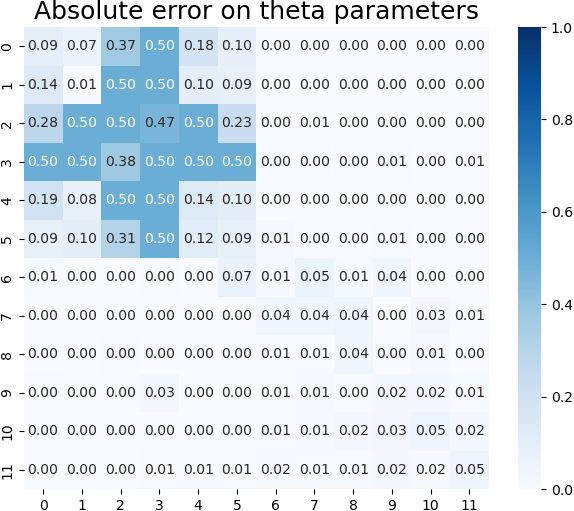}
     \caption{}
\end{subfigure}
\caption{Learned $\theta_{ij}$ parameters (a) and Absolute Error (b) for misconfigured $\pta$}
\label{fig: misconfigured pta 1}
\endminipage\hfill
\minipage{0.49\textwidth}
\begin{subfigure}[b]{0.49\textwidth}
     \centering
     \includegraphics[width=\linewidth]{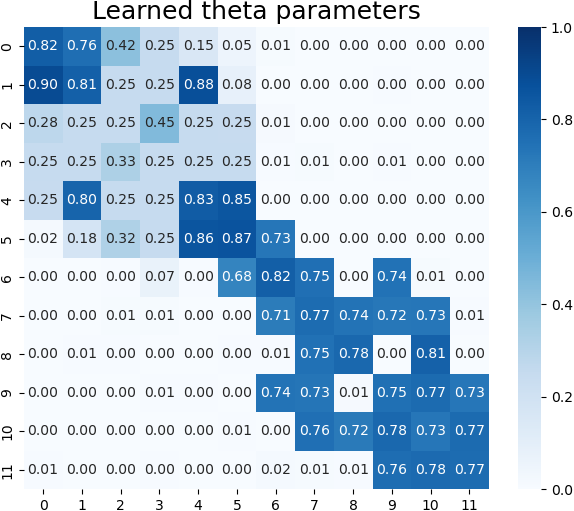}
     \caption{}
\end{subfigure}
\hfill
\begin{subfigure}[b]{0.49\textwidth}
     \centering
     \includegraphics[width=\linewidth]{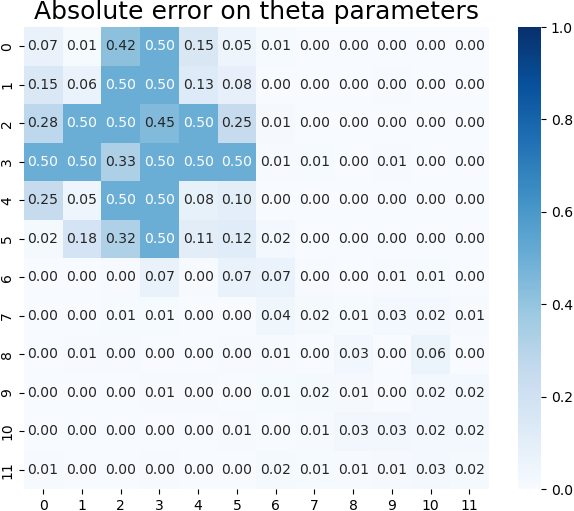}
     \caption{}
\end{subfigure}
\caption{Learned $\theta_{ij}$ parameters (a) and Absolute Error (b) for misconfigured $\pta$}
\label{fig: misconfigured pta 2}
\endminipage\hfill
\end{figure}

\paragraph{Misconfigured $\pta$} \label{par: misconfigured P}
Figures \ref{fig: misconfigured pta 1} and \ref{fig: misconfigured pta 2} correspond to the experiment where some $\theta_{ij}$ values of $\pta$ are fixed at incorrect values, while the processing function $f_\psi$ is fixed to the true one. In the community affected by the perturbation, free $\theta_{ij}$ values tend to be sampled more frequently to compensate for the downsampling imposed by the perturbation. Interestingly, all the edges with at least one edge in the second community ($75 \%$ of the edges) appear unaffected by the perturbation.

\newpage

\subsection{Additional details on the experiments of Section \ref{sec: comparison with literature}} \label{appendix: comparison with literature}

All models have been learned using SFE. 
$\lossfunc$ rely on the following gradient with respect to $\theta$
$$
\nabla_\theta \lossfunc 
= \E_{x,y^*}[2(\E_A[\hat y] - y^*) \E_A[\hat y \nabla_\theta \log\pta(A)]].
$$
For $\losslit_1$, the gradient is rewritten as
$$
\nabla_\theta \losslit_{1,\ell} 
=   \E_{x,y^*,A}[(\ell(\hat y, y^*)-b)\nabla_\theta\log\pta(A)]
$$
with $b$ estimating the expected value $\E_{x,y^*,A}[\ell(\hat y, y^*)]$.

The second family of loss functions $\losslit_2$ focuses on node-level prediction rewriting $\ell(\hat y, y^*)$  as the mean $\tfrac{1}{N}\sum_{i=1}^N\ell(\hat y_i, y^*_i)$ over the prediction error at each node $i$. The gradient with respect to $\theta$ is then written as
\begin{align}\label{eq: kazi loss}
\nabla_\theta \losslit_{2,\ell} &= \E_{x, y^*} \E_{A \sim \pta} \left[\frac{\sum_{i}^{N}(\ell(y_i, y^*_i) - b_i)\nabla_\theta \log(\pta(A_{i,:}))}{N}\right] 
\end{align}
where $b_i$ are computed as moving averages of $\ell(y_i, y^*_i)$.

The last family of loss functions (i.e., $\loss^{\text{literature}}_{\text{elbo}}$) requires (i) prior distributions $\bar{P}_A(A)$ and $(ii)$ a standard deviations for $P^\psi_{y|x^*,A}(y^*)$ to be set. We consider the following priors and standard deviations, selecting the combination with the lowest validation loss:

\begin{enumerate}[label={(\roman*)}]
\item
For the prior distributions, we assume that each edge is sampled independently according to a Bernoulli distribution. We consider three different prior specifications:
\begin{itemize}
\item The first prior is a Bernoulli distribution with parameter $p = 0.01$ for all edges.
\item The second prior is a Bernoulli distribution with parameter $p = 0.5$ for all edges.
\item The third prior is defined based on the ground truth graph structure: for edges sometimes present in the ground truth structure (i.e., $\theta^*_{ij} \ne 0$), the Bernoulli parameter is $p = 0.75$, while for edges never present in the ground truth structure (i.e., $\theta^*_{ij} = 0$), the parameter is $p = 0.05$.
\end{itemize}
\item
the standard deviations considered are: \{$0.001$, $0.005$, $0.01$, $0.05$, $0.1$, $0.5$\}.
\end{enumerate}

\subsection{Real-world experiment}\label{appendix: AQI experiment}
\begin{figure}[H]
    \centering
    \includegraphics[width=0.55\linewidth]{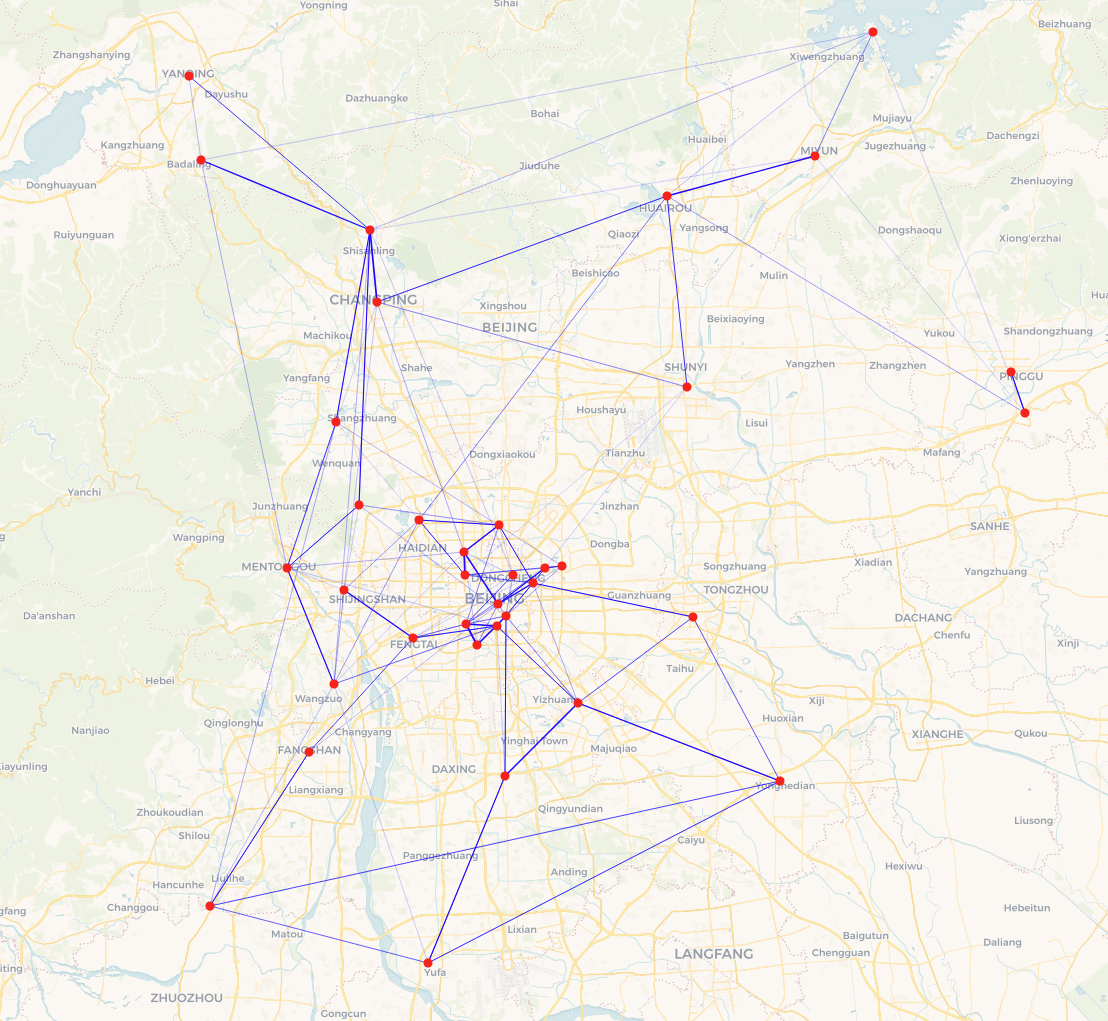}
    \caption{Graph structure learned by our approach on Beijing air quality data. The nodes correspond to sensor locations. The thickness of the edges is proportional to the corresponding probability. Map data from \hyperlink{https://www.openstreetmap.org/copyright}{OpenStreetMap}.}
    \label{fig: AQI}
\end{figure}

To demonstrate that our method learns meaningful graph distributions in real-world settings, we train a neural network on air quality data in Beijing \cite{zheng2013u}. The dataset consists of pollutant measurements collected by sensors in Chinese urban areas across several months. We do not use this dataset as a benchmark because the graph structure provided with the data is based on physical distance, and there is no guarantee that it represents the true underlying structure. The neural network we use consists of a GRU unit for processing each time series, followed by a GNN with a learnable graph structure. Figure \ref{fig: AQI} shows the graph structure learned by our approach, demonstrating its capability to learn meaningful distributions in real-world settings.

\subsection{Compute resources and open-source software}
\label{appendix: Compute resources and Libraries used}

The paper's experiments were run on a workstation with AMD EPYC 7513 processors and NVIDIA RTX A5000 GPUs; on average, a single model training terminates in a few minutes with a memory usage of about 1GB. 

The developed code relies on PyTorch~\cite{paszke2019pytorch} and the following additional open-source libraries: 
PyTorch Geometric~\cite{fey2019fast}, NumPy~\cite{harris2020array} and Matplotlib~\cite{hunter2007matplotlib}.

\end{document}